\documentclass[journal]{IEEEtran}
\usepackage{cite}
\usepackage{amsmath,amssymb,amsfonts}
\usepackage{graphicx}
\usepackage{caption}
\usepackage{subfig}

\usepackage{algorithm}      
\usepackage{algpseudocode}
\usepackage{multirow}
\usepackage{multicol}
\usepackage{threeparttable}
\usepackage{textcomp}
 \usepackage{booktabs}
\def\BibTeX{{\rm B\kern-.05em{\sc i\kern-.025em b}\kern-.08em
    T\kern-.1667em\lower.7ex\hbox{E}\kern-.125emX}}
\usepackage{enumitem}

\begin{document}

\title{Real-Time Air Pollution Monitoring and Tracking via Adaptive Density-Driven Multi-UAV Coordination}
\author{Julian~Martinez,
        Ruchika~Singh,
        Mohammad~Afrazi,
        and~Kooktae~Lee
\thanks{*This work was supported by NSF CAREER Grant CMMI-DCSD-2638508.}
\thanks{J. Martinez, R. Singh, and K. Lee are with the Department of Mechanical and Aerospace Engineering, Texas Tech University, Lubbock, TX 79409 USA (e-mail: mar85781@ttu.edu; ruchisin@ttu.edu; kooktae.lee@ttu.edu).}%
\thanks{M. Afrazi is with the Department of Mechanical Engineering and Materials Science, Duke University, Durham, NC 27708 USA (e-mail: mohammad.afrazi@duke.edu).}}

\maketitle

\begin{abstract}
Accurate and timely air pollution monitoring and tracking are critical for protecting public health and supporting effective environmental management. This paper presents an adaptive Density-Driven Control (D$^2$C) framework for coordinating multiple unmanned aerial vehicles (UAVs) to track and map dynamically evolving pollutant distributions. The proposed framework extends the D$^2$C strategy to guide UAV trajectories based on predictive pollutant propagation and dynamically updated reference densities. A sensor-driven surrogate plume update mechanism is employed, in which reference samples are propagated using locally sensed concentration gradients and wind estimates through gradient attraction, wind advection, inter-sample repulsion, and stochastic diffusion, enabling UAVs to maintain an accurate spatio-temporal representation of the pollution field. The framework utilizes a coordinated structure where individual UAVs share their local measurements with a central coordinator to maintain a unified set of reference samples, facilitating systematic target allocation and trajectory planning. Simulation results under both mild and strong wind conditions demonstrate that the proposed method significantly outperforms state-of-the-art baselines in terms of Sinkhorn distance. Specifically, in scenarios where the plume drifts beyond a fixed monitoring domain under high-velocity, time-varying wind fields, the tested baseline methods fail to maintain tracking, whereas the proposed D$^2$C framework successfully adapts to the evolving distribution. These findings highlight D$^2$C as an efficient and resilient solution for atmospheric monitoring in dynamic environments.
\end{abstract}

\section{Introduction}
\label{sec:introduction}

Air pollution poses significant threats to human health and the environment \cite{manisalidis2020environmental}. The World Health Organization (WHO) estimates that it causes approximately seven million premature deaths annually \cite{manisalidis2020environmental}. Major pollutants include particulate matter (PM), nitrogen oxides (NOx), sulfur dioxide (SO$_2$), volatile organic compounds (VOCs), and carbon dioxide (CO$_2$). These substances are associated with respiratory and cardiovascular diseases, as well as acid rain, photochemical smog, and climate change.

Ground-based stations are commonly used to assess air quality in urban and industrial areas. Although ground-based stations provide continuous local measurements, their fixed deployment restricts the area that can be observed \cite{saini2025robot, carotenuto2023low}. The problem becomes more pronounced as wind, temperature inversions, and atmospheric turbulence alter the direction and rate of pollutant dispersion \cite{sorek2020strategies}. As a result, the movement of pollution plumes may not be adequately captured by measurements from stationary networks.

Satellite remote sensing offers a means of extending atmospheric observations beyond the locations covered by ground stations. Instruments such as MODIS and Sentinel provide measurements across extensive geographical regions \cite{holloway2021satellite, kaplan2020space}. However, their spatial resolution and observation intervals restrict the detection of localized and rapidly changing pollution events. In addition, cloud cover and atmospheric interference can lead to gaps in the collected data \cite{stratoulias2024recent, holloway2021satellite}. Thus, satellite observations provide broad coverage but remain limited in their ability to follow plume evolution at short time intervals.

UAVs equipped with environmental sensors can obtain pollutant measurements at locations that are not covered by fixed monitoring stations \cite{kabir2021wildlife, kokate2023review, jaron2024assessment, liang2024integrated, facinelli2019cooperative, sharma2025unmanned}. Deploying several UAVs over the same region makes it possible to examine the pollution field from different locations. In this context, \textit{pollution monitoring} refers to estimating the spatial distribution of pollutant concentrations, whereas \textit{pollution tracking} concerns the temporal changes in plume position and distribution. Both functions are important when investigating pollution phenomena that vary across space and time.

A major issue, however, lies in the way measurement routes are assigned. Many existing coordination schemes prescribe the flight paths in advance, commonly using lawnmower or grid-based patterns \cite{galceran2013survey, gaalfalk2021sensitive}. Once assigned, these paths provide little flexibility for responding to variations in pollutant levels or plume movement. UAVs may therefore spend measurement time in regions that no longer reflect the areas of interest. Allowing flight paths to change with the observed data is consequently an important consideration in multi-UAV coordination \cite{kabir2021wildlife}.

To address these challenges, this paper extends the Density-Driven Control (D$^2$C) framework \cite{kabir2021wildlife, lee2022density, afrazi2025density, afrazi2025enhanced} to enable adaptive multi-UAV coordination for real-time air pollution \textit{monitoring and tracking}. Although the Density-Driven Optimal Control (D$^2$OC) further improved the performance of D$^2$C by considering the optimality, they are mostly limited to the linear time-invariant (LTI) dynamics \cite{seo2025tcst, seo2025density, lee2025connectivity, lee2026tac, lee2026automatica,lee2026ai}. Thus, D$^2$OC 
may not be directly applicable to air pollution monitoring missions, where the mobile platforms such as UAVs have nonlinear dynamics.

The proposed approach leverages Optimal Transport (OT) theory \cite{villani2009optimal} to dynamically guide UAVs toward regions having higher pollutant densities, effectively coupling predictive propagation with data-driven adaptation. The framework maintains a dynamically evolving reference density by updating reference samples with local concentration gradients and wind measurements. This enables UAV trajectories to adapt seamlessly to changes in the pollutant distribution.
To achieve this in practice, each UAV uses onboard gas and particulate sensors to obtain localized, point-wise pollutant measurements \cite{rohi2020autonomous,al2023bayesian}, together with local wind information. 
Each UAV collects local wind and concentration-gradient measurements and sends them to a central coordinator. The coordinator uses these inputs to update a common reference distribution. Its evolution is governed by four components: gradient attraction, wind advection, inter-sample repulsion, and stochastic diffusion. The updated density distribution determines the targets for a separate motion-control stage, which generates trajectories while accounting for UAV dynamics.

We evaluate the proposed centralized method against two alternatives: a prediction-only propagation model and a distributed ergodic coverage approach using Gaussian-process field estimation \cite{mantovani2026distributed}. The latter coordinates multiple agents by matching the time-averaged spatial distribution of their trajectories to a time-varying target distribution.
Comparing against these baselines allows us to isolate and evaluate the distinct performance benefits of our measurement-informed reference updates and centralized coordination over open-loop predictions and distributed ergodic alternatives.

The main contributions of this work are threefold:
1) A centralized density-driven coordination framework for multi-UAV pollution monitoring and tracking, in which a common time-varying reference distribution is used to coordinate individual UAV targets and allocate coverage weights across the monitored region. This formulation enables multiple UAVs to distribute their sensing activities according to the spatial characteristics of the pollution field;
2) A measurement-informed reference update mechanism that incorporates locally sensed concentration gradients and wind measurements into the evolution of the shared reference distribution. By accounting for gradient attraction, wind-driven advection, inter-sample repulsion, and stochastic diffusion, the proposed mechanism adjusts the reference samples according to changes in plume distribution and provides updated targets for UAV motion control;
3) A numerical evaluation of the integrated sensing, reference-update, and trajectory-control framework under spatially varying wind conditions. The proposed centralized approach is compared with a prediction-only propagation model and a distributed ergodic coverage method based on Gaussian-process field estimation, allowing the effects of measurement-based adaptation and centralized coordination on pollution monitoring and tracking performance to be examined.

The remainder of the paper is organized as follows. Section II introduces the D$^2$C formulation and the centralized monitoring problem. Section III describes the pollutant propagation model, Section IV develops UAV sensing and reference-sample propagation, and Section V presents the integrated multi-UAV algorithm. Simulation results are discussed in Section VI, followed by conclusions in Section VII.

\section{PRELIMINARY AND PROBLEM DESCRIPTION}
\label{sec:preliminaries}

\noindent\textbf{Notations:} 
We denote the set of real numbers by $\mathbb{R}$. A real-valued vector with $n$ components is represented by $\mathbb{R}^n$, and $\mathbb{R}^{n \times m}$ denotes a matrix with $n$ rows and $m$ columns of real numbers. The identity matrix of dimension $n \times n$ is represented as $\mathbf{I}_n$. The $\ell_2$-norm (Euclidean norm) of a vector $x$ is denoted by $\|x\|$. 
The cardinality of a set $\mathcal{S}$, i.e., the number of elements in $\mathcal{S}$, is denoted by $|\mathcal{S}|$. 
A sequence of elements indexed from $1$ to $N$ is indicated as $\{\mathbf{q}_j\}_{j=1}^{N} = \{q_1, q_2, \dots, q_N\}$ and may be simplified to $\{\mathbf{q}_j\}$ when indexing limits are clear from context. 
A real-valued random variable $x$ following a normal distribution with mean $\mu$ and variance $\sigma^2$ is denoted as $x \sim \mathcal{N}(\mu, \sigma^2)$. 

\medskip

The D$^2$C framework \cite{kabir2021wildlife} coordinates a group of robots to achieve a desired spatial distribution through their time-averaged trajectories. For air quality monitoring, this distribution can be designed to assign greater sampling priority to areas with elevated pollution levels. Its construction may incorporate satellite observations, meteorological forecasts, industrial emission records, and ground-based measurements. While satellite data provide information on regional pollutant dispersion, emission inventories offer insight into their possible origins. Considering both sources in the reference density allows the identification of relevant regions and the allocation of UAV sampling accordingly.

The use of UAVs also changes how environmental measurements are collected compared with conventional fixed ground monitoring systems \cite{yang2017real, bakirci2024enhancing}. Rather than relying exclusively on measurements from predetermined locations, mobile platforms can obtain observations at different positions as pollution conditions evolve. Incorporating updated environmental information into D$^2$C allows the spatial allocation of sensing activity to change accordingly. This provides a basis for improving the coverage of the monitored area and examining both pollutant distributions and plume movement.

In the following subsections, we present the D$^2$C formulation, describe its implementation, and provide an overview of the problem.

\subsection{Density-Driven Controller (D$^2$C)}

\begin{figure}[!t]
    \centering
    \includegraphics[width=0.7\columnwidth]{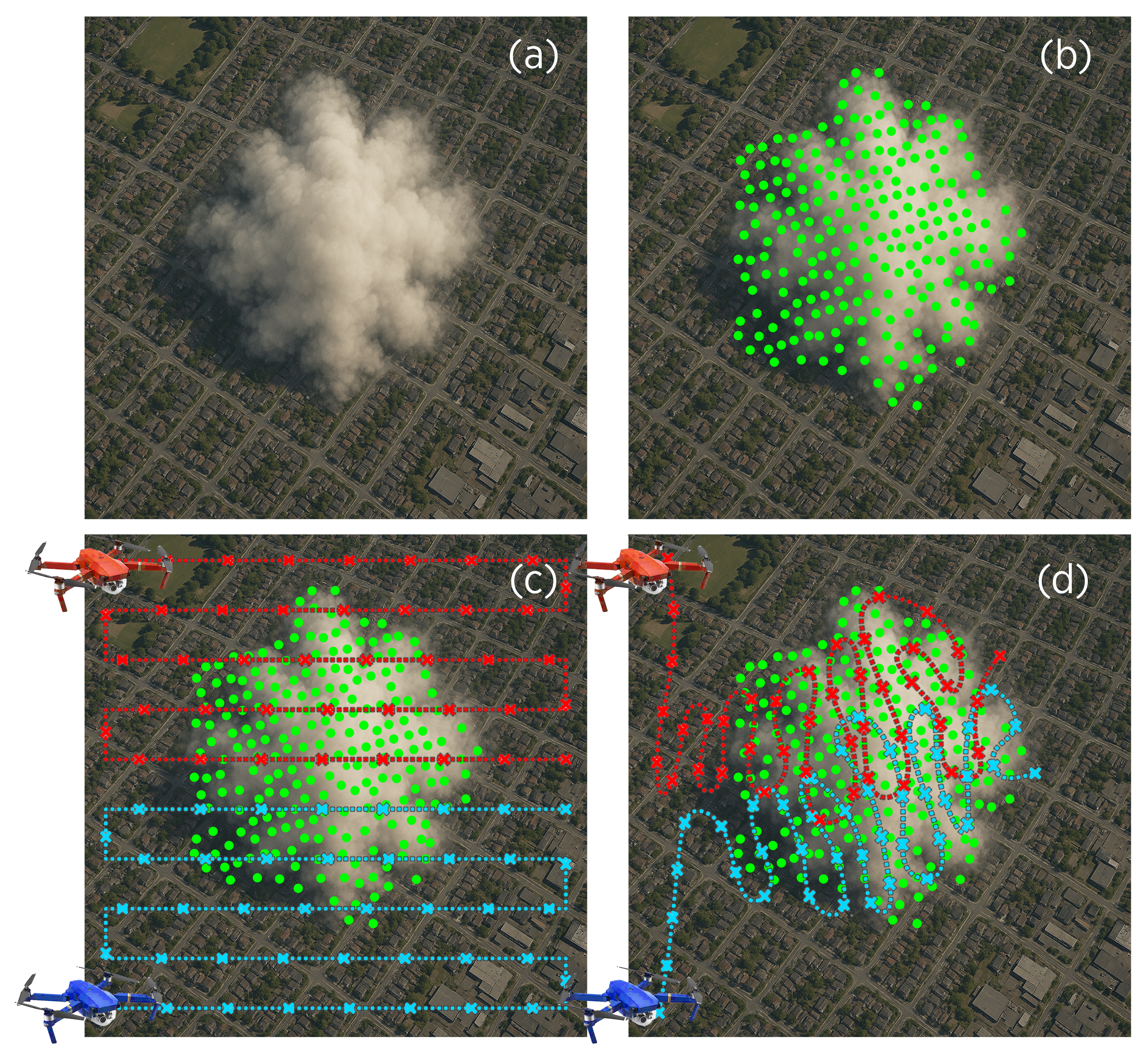}
    \caption{Illustration of UAV-based area monitoring: a) Satellite imagery of urban area; b) Sample-point representation of a reference pollution density distribution; c) Conventional area coverage path (LM path); d) $D^{2}C$ method aligning agents with high-priority zones.}
    \label{fig:D$^2$C}
\end{figure}

In Figure~\ref{fig:D$^2$C}, two different pollutant monitoring strategies are illustrated. Figure~\ref{fig:D$^2$C}(a) shows an illustrative pollutant plume over an aerial image of the monitoring area, and Figure~\ref{fig:D$^2$C}(b) shows its representation by reference sample points (SPs). In Figure~\ref{fig:D$^2$C}(c), conventional lawnmower (LM) coverage is depicted, where the path is generated to follow predefined parallel sweeps regardless of pollutant intensity. In contrast, Figure~\ref{fig:D$^2$C}(d) describes density-driven UAV paths that focus sensing effort on regions with higher reference-sample density. In the D$^2$C formulation, a common planning process uses the reference distribution to coordinate target selection across UAVs. In the following, the formal process for D$^2$C as well as the OT theory used as a tool to realize D$^2$C are introduced.

\vspace{1mm}

\subsubsection{Optimal Transport (OT) Theory}
\begin{figure}[t]
  \centering
  \includegraphics[width=.82\columnwidth]{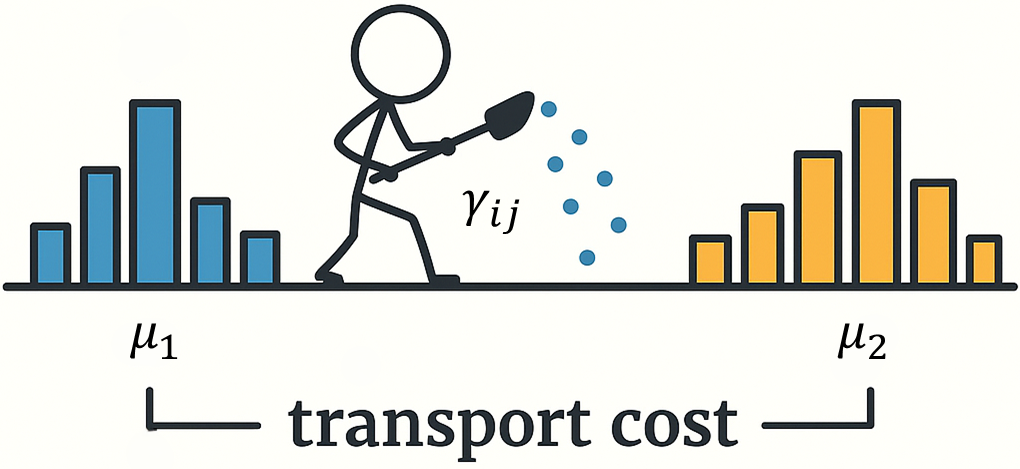}
  \caption{Optimal Transport principle for matching two weighted distributions.}
  \label{fig:OTtheory}
\end{figure}

Optimal Transport provides a measure of the cost of matching two probability distributions \cite{villani2009optimal,chen2021optimal} (Figure~\ref{fig:OTtheory}). Let $\mu_1$ and $\mu_2$ be discrete distributions supported on the agent points (APs) $\{\mathbf p_i\}_{i=1}^{M}$ with weight $\alpha_i\geq0$ and reference sample points $\{\mathbf q_j\}_{j=1}^{N}$ with weight $\beta_j\geq0$. Their squared 2-Wasserstein distance is defined by
\begin{equation}
W_2^2(\mu_1,\mu_2)=\min_{\gamma\geq 0}\sum_{i=1}^{M}\sum_{j=1}^{N}\gamma_{ij}\|\mathbf p_i-\mathbf q_j\|^2,
\label{equation_Wasserstein_distance}
\end{equation}
subject to
$\sum_{j=1}^{N}\gamma_{ij}=\alpha_i,\, \forall i$, 
$\sum_{i=1}^{M}\gamma_{ij}=\beta_j,\, \forall j$, and
$\sum_{i=1}^{M}\alpha_i=\sum_{j=1}^{N}\beta_j=1$,
where $\gamma_{ij}$ denotes the mass transport plan between AP $i$ and SP $j$. The first two equalities represent the mass assignment constraints for $\gamma_{ij}$ while the last equality corresponds to the mass conservation law.

Although equation \eqref{equation_Wasserstein_distance} describes the standard full-mass OT problem, D$^2$C instead uses the remaining SP weights for a \emph{partial, incremental} coverage update. Thus, its residual weights need not sum to one after each step.

\subsubsection{Agent and Reference Weights}
In the centralized setting, the reference samples and their remaining weights are shared by all UAVs. At initialization, equal SP weights are assigned to each SP as follows:
\begin{equation}
\beta_j^0=\frac{1}{N_{\text{REF}}},\qquad
\sum_{j=1}^{N_{\text{REF}}}\beta_j^0=1,
\label{eq:beta_init}
\end{equation}
where $N=N_{\text{REF}}$ is the total number of reference SPs.

A higher density in a region, represented by more SPs, indicates greater monitoring priority even when the initial weights are uniform. Each UAV contributes an incremental mass budget $\alpha_r^k$ at step $k$. For  identical, $R$ numbers of UAVs with a common coverage horizon $T_{\mathrm{total}}$ and a discrete-time sampling interval $\Delta t$, the team-normalized weight for each UAV is
\begin{equation*}
\alpha_r^k=\frac{\Delta t}{R\,T_{\mathrm{total}}},\qquad r\in R,
\end{equation*}
imposing the condition that the weight for each AP is uniform. This is a \emph{team-level} normalization: the total planned contribution across all UAVs and all update steps is one. Conceptually, at each time step, some weights of SPs, $\beta_j^k$, are transported to those of APs, $\alpha_r^k$, reducing $\beta_j^k$ over time for the coverage progress. In this way, a multi-UAV team can cover the domain according to the predefined reference distribution. More details about the D$^2$C procedure are explained below.

\subsubsection{Three-Stage Centralized Control Procedure}
The controller consists of (A) target SP selection, (B) UAV motion control, and (C) centralized reference-weight updating. Local measurements from each UAV using onboard sensors are supplied to the common reference-update process described in Section IV.

\begin{enumerate}[label=\roman*), leftmargin=*, labelsep=5pt]
\item Stage A: Centralized Target Selection. At each planning step, each UAV $r$ greedily selects the closest SPs from the active set $\mathcal{Q}_+^k = \{\mathbf{q}_j^k : \beta_j^k > 0\}$ in order of increasing distance until their cumulative weight exactly matches the budget $\alpha_r^k$. The resulting subset $\mathcal{Q}_{r}^k$ determines the assigned target via its weighted centroid,
\begin{equation}
\mathbf{q}_{\mathrm{goal},r}^k = \frac{\sum_{\mathbf{q}_j^k \in \mathcal{Q}_{r}^k} \beta_j^k \mathbf{q}_j^k}{\sum_{\mathbf{q}_j^k \in \mathcal{Q}_{r}^k} \beta_j^k}.
\label{eq:centroid}
\end{equation}
To prevent independent double consumption of weights across UAVs, allocations are coordinated sequentially by the predefined agent order. 
While the original D$^2$C \cite{kabir2021wildlife,lee2022density} relies on an exhaustive permutation search among nearby SPs to select the next target, the proposed centroid rule above avoids this heavy computational burden.

\item Stage B: UAV Motion Control. After centralized target assignment, each UAV moves toward its target using an appropriate vehicle-level controller. For a discrete-time nonlinear model $f$:
\begin{equation*}
\mathbf x_r^{k+1}=f(\mathbf x_r^k, \mathbf u_r^k),
\end{equation*}
a feedback law of the form $\mathbf u_r^k=K_r(\mathbf x_{d,r}^k-\mathbf x_r^k)$ may be used with some gain $K_r$ and the desired state $\mathbf x_{d,r}^k$ containing the target position $\mathbf q_{\mathrm{goal},r}^k$ when the gain and target are chosen to satisfy the corresponding tracking and stability conditions. Alternatively, more advanced controller $\mathbf u_r^k=g_r(\mathbf x_r^k,\mathbf x_{d,r}^k)$ such as an MPC may instead be considered. 
More detailed description regarding the controller used in our study is explained in the simulation,  Section VI. Although we adopt a centralized communication for the unified pollutant information update, each agent can independently compute its control law in a distributed fashion.

\item Stage C: Centralized Reference-Weight Updating. Following the UAV motion update, the common reference weights are reduced according to the coverage mass allocated to the entire team,
\begin{equation*}
\beta_j^{k+1}=\beta_j^k-\sum_{r=1}^{R}\gamma_{rj}^{k,*},
\end{equation*}
where $\gamma_{rj}^{k,*}$ stands for the optimal mass transport plan, whose solution can be obtained analytically according to \cite{lee2022density}.

One centralized mass-allocation formulation consistent with the shared-weight representation is
{
\begin{equation}
\begin{aligned}
\min_{\{\gamma_{rj}^k\geq 0\}}\quad &
\sum_{r=1}^{R}\sum_{j=1}^{N_{\text{REF}}}
\gamma_{rj}^k\|\mathbf {}^r\mathbf{p}^{k+1}-\mathbf q_j^k\|^2\\
\text{subject to}\, &
\sum_{j=1}^{N_{\text{REF}}}\gamma_{rj}^k=\alpha_r^k,\, \forall r,
\sum_{r=1}^{R}\gamma_{rj}^k=\beta_j^k,\, \forall j.
\end{aligned}
\label{eq:central_lp}
\end{equation}}
\end{enumerate}

\begin{figure}[t]
  \centering
  \includegraphics[width=.83\columnwidth]{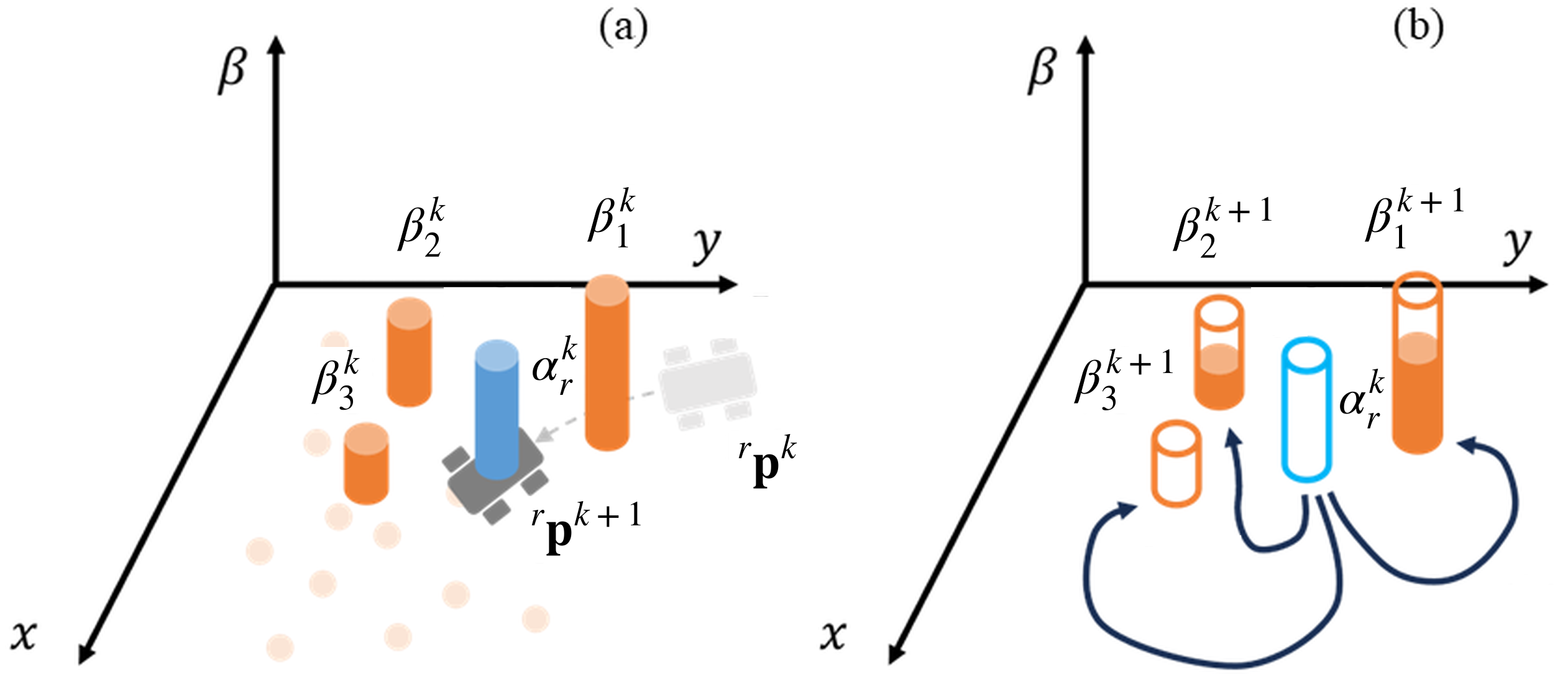}
    \caption{Reference sample weights before (left) and after (right) the weight update. Each cylinder is a reference sample $j$ with weight $\beta_j^k$. UAV $r$ moves from $^r\mathbf{p}^k$ to $^r\mathbf{p}^{k+1}$ and removes a total weight $\alpha_r^k$ from the samples it covers, which gives the updated weights $\beta_j^{k+1}$.}  \label{fig:weigh_update}
\end{figure}

Note that the above equality constraints are for the mass allocation condition, same as the original OT problem. An illustration is provided in Figure~\ref{fig:weigh_update} to explain the reference-weight update concept.

\subsection{Problem Overview}
\label{sec:problemoverview}
The core challenge addressed by this research is the robust detection, monitoring, and tracking of air pollution sources and their dispersion patterns relying solely on initial satellite observations. Addressing these spatio-temporal limitations and operational challenges requires innovative methods that integrate multiple data sources and dynamically adapt to partial data.

To overcome these challenges, we propose a multi-UAV system guided by the D$^2$C framework \cite{kabir2021wildlife}. Let $c(x,t)$ denote an unknown, time-varying pollutant concentration field over a monitored three-dimensional region $\Omega \subseteq \mathbb{R}^{3}$. UAV r obtains measurements at its current location, including a concentration-gradient estimate and a local wind measurement. The centralized process uses these observations to maintain a
common set of reference samples and assign monitoring targets under the UAVs'
motion constraints. It is worth noting that the true plume distribution is completely unknown to the UAV team. The control objective is to track the evolving pollutant field while directing collective sensing effort toward regions represented by the updated reference density. Thus, the controller must (i) update its surrogate plume from incomplete, noisy measurements; (ii) assign sensing targets to the UAV team
using a common reference distribution; and (iii) account consistently for the coverage mass assigned across UAVs. For evaluation, we compare the resulting reference distribution with the ground-truth plume using prediction-only propagation and the ergodic coverage benchmarks. Next, we detail the wind model, plume propagation, sensing, and integrated algorithm.

\section{Simulation Framework for Wind-Driven Plume Propagation}
\label{sec:propagation}

Simulating realistic pollution transport requires capturing multiple interacting atmospheric phenomena rather than relying on idealized uniform flows. To test our multi-UAV coordination strategy under harsh operational conditions, we build a simulator where pollutant particles evolve in a three-dimensional domain $\mathbf{q} = [x, y, z]^T \in \mathbb{R}^3$. In this environment, wind fields vary across space, which forces UAVs to measure local conditions directly rather than relying on precition models.

\subsection{Modeling Spatial Wind Gradients}
\label{subsec:local_wind}

Real wind fields shift across terrain. While allowing height govern vertical variations, we restrict dominant advection to the horizontal plane as follows:
\begin{equation}
    \mathbf{w}(\mathbf{q}) = 
    \begin{bmatrix}
        \cos \theta_{\text{wind}}(\mathbf{q}) \\
        \sin \theta_{\text{wind}}(\mathbf{q})
    \end{bmatrix} \in \mathbb{R}^2,
\end{equation}
where $\theta_{\text{wind}}(\mathbf{q})$ depends on the particle position $\mathbf{q}$. We divide the workspace into indexed regions $m$. Each region gets its own baseline wind angle $\theta_{\text{wind},0}^{(m)}$, noise level, and directional drift. As pollutant particles propagate across regions, they experience abrupt and continuous changes in wind direction and magnitude. Global wind assumptions break down completely under this spatial patchiness, leading to the necessity of onboard UAV wind sensing for tracking.

\subsection{Advection, Drift, and Shear Mechanics}
\label{subsec:wind_drift_shear}

Primary transport follows local wind advection:
\begin{equation}
    \mathbf{q}_{\text{adv}} = v_{\text{wind}}(\mathbf{q}) \, \tilde{\mathbf{w}}(\mathbf{q}) \, \Delta t,
\end{equation}
where $v_{\text{wind}}(\mathbf{q})$ is local wind speed, and $\tilde{\mathbf{w}}(\mathbf{q}) = [\mathbf{w}(\mathbf{q})^T,\ w_z]^T \in \mathbb{R}^3$ maps the 2D vector into 3D space. Slow directional wandering inside each zone follows:
\begin{equation}
    \theta_{\text{wind}}(\mathbf{q}) = \theta_{\text{wind},0}(\mathbf{q}) + \theta_{\text{bias}}(\mathbf{q}) + \xi_{\theta_{\text{wind}}},
\end{equation}
with deterministic drift $\theta_{\text{bias}}(\mathbf{q})$ and Gaussian noise $\xi_{\theta_{\text{wind}}} \sim \mathcal{N}(0,\sigma_{\theta_{\text{wind}}}^2)$. 

Furthermore, horizontal wind speed increases with altitude. We model this vertical wind shear via linear height scaling:
\begin{equation}
    \mathbf{q}_{\text{shear}} = v_{\text{wind}}(\mathbf{q}) \, \kappa_s \frac{z}{z_{\max}}
    \begin{bmatrix} 1 \\ 1 \\ 0 \end{bmatrix} \Delta t,
\end{equation}
where $\kappa_s$ is the shear coefficient and $z_{\max}$ normalizes altitude. Since shear is zero on the ground, higher particles move faster along different headings.

\subsection{Diffusion, Gusts, and Vortex Dynamics}
\label{subsec:diffusion_vortex}

Small-scale mixing is modeled via anisotropic turbulent diffusion aligned with local streamlines:
\begin{equation}
    \mathbf{q}_{\text{turb}} = \tau 
    \left[
    (\mathbf{r}\cdot\tilde{\mathbf{w}}(\mathbf{q}))\,\tilde{\mathbf{w}}(\mathbf{q}) 
    + 
    \kappa\left(\mathbf{r} - (\mathbf{r}\cdot\tilde{\mathbf{w}}(\mathbf{q}))\,\tilde{\mathbf{w}}(\mathbf{q})\right)
    \right]
    \sqrt{\Delta t},
\end{equation}
where $\mathbf{r} \sim \mathcal{N}(\mathbf{0},\mathbf{I}_3)$, turbulence intensity is $\tau$, and cross-wind spread is scaled by $\kappa$.

Since standard Gaussian noise cannot replicate rare, long-range jumps, which exist in reality, L\'evy flight perturbations are added:
\begin{equation}
    \mathbf{q}_{\text{Levy}} = \ell_{\text{scale}}\, \mathbf{L}_\alpha,
\end{equation}
where $\mathbf{L}_\alpha \in \mathbb{R}^3$ comes from a truncated symmetric Cauchy distribution scaled by $\ell_{\text{scale}}$. 
This motion captures violent thermal updrafts and heavy gusts to reflect L\'evy flight perturbations.

Finally, localized swirling is produced by rotational velocity fields around centers $\mathbf{c}_i \in \mathbb{R}^3$:
\begin{equation}
    \mathbf{q}_{\text{vortex}} = 
    \sum_{i=1}^{N_v} \mathbf{vor}_i(\mathbf{q}-\mathbf{c}_i)\,\Delta t,
\end{equation}
where $N_v$ counts active vortices and $\mathbf{vor}_i(\cdot)$ dictates decay with distance.

\subsection{Integrated Propagation and Tracking Rationale}
\label{subsec:combined_tracking}

Summing all six components provides the complete position update rule for each particle as follows:
\begin{equation}
\label{eq:combined_model}
    \mathbf{q}^{k+1} = \mathbf{q}^{k}
    + \mathbf{q}_{\text{adv}}
    + \mathbf{q}_{\text{shear}}
    + \mathbf{q}_{\text{turb}}
    + \mathbf{q}_{\text{Levy}}
    + \mathbf{q}_{\text{vortex}}.
\end{equation}

\begin{figure}[!h]
    \centering
    \subfloat[]{
    \includegraphics[width=0.5\linewidth]{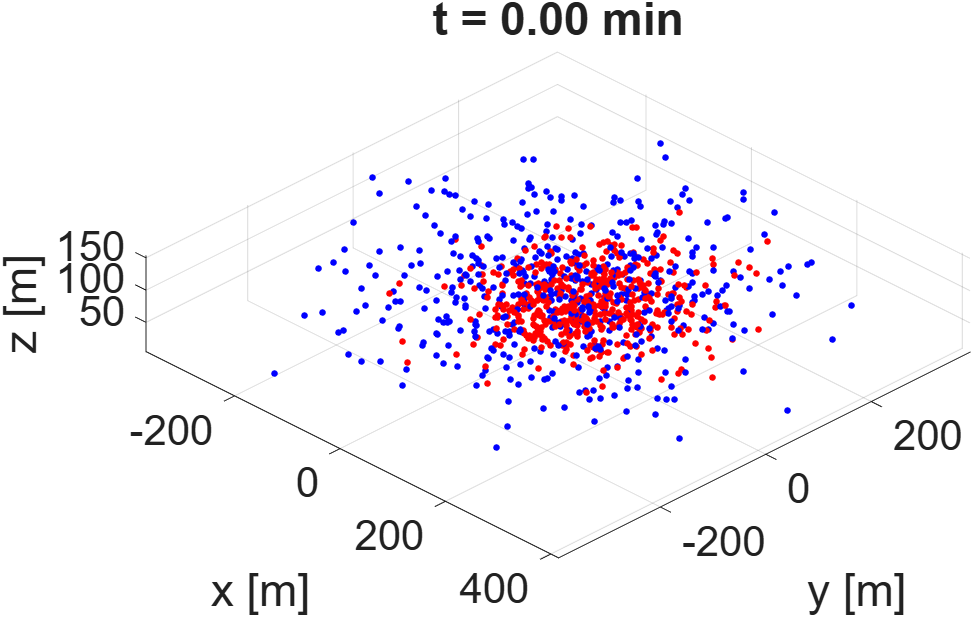}}
    \subfloat[]{
    \includegraphics[width=0.5\linewidth]{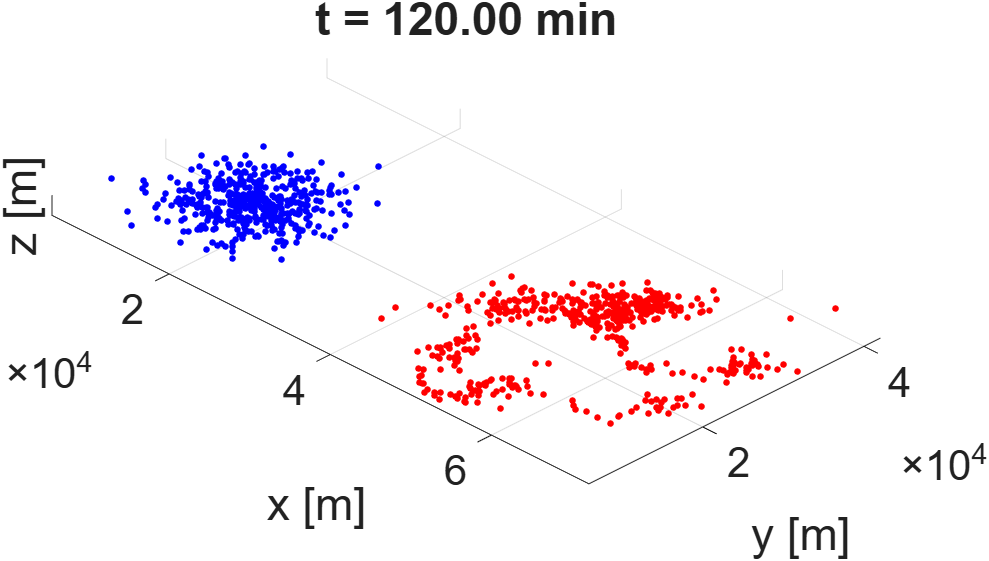}}
    \caption{Propagation of pollutant particles over a 2-hour period using the combined dynamic model. Blue particles correspond to nominal wind parameters, while red particles correspond to ground-truth wind parameters differing from the nominal values. Initially Gaussian-distributed particles evolve under advection, shear, vortex, turbulence, and Lévy flight effects.}
    \label{fig:propagation}
\end{figure}

Using this model, we simulated the initially Gaussian-distributed pollutant particles over a 2-hour period, as illustrated in Figure~\ref{fig:propagation}. Blue particles correspond to propagation using nominal wind parameters, while red particles correspond to propagation using ground-truth wind parameters that differ from the nominal values.

\section{UAV-Based Pollutant Sensing and Local Gradient Estimation}
\label{sec:posterior_update}

This section presents the UAV-based pollutant sensing framework used in our simulations. Each UAV is treated as an autonomous sensing agent that measures local wind velocity and estimates pollutant concentration based on surrounding plume particles. The UAV then computes local concentration gradients, which are used to guide motion within the D$^2$C framework. Each UAV computes its gradient and wind estimates from its own onboard measurements only. No UAV has access to the ground-truth plume state.

\subsection{Onboard Sensors and Measurement Model}

The position of each UAV $r$ at discrete time step $k$ is given by
\begin{equation}
    {}^r \mathbf{p}^k = [x^k, y^k, z^k]^\top \in \mathbb{R}^3.
\end{equation}

The UAV is assumed to measure the following physical quantities using onboard sensors.

\subsubsection{Local wind velocity:} The UAV senses the wind velocity vector at its position, denoted as
    \begin{equation}
        \mathbf{u}_{\text{wind}}({}^r \mathbf{p}^k) = 
        \begin{bmatrix} u_x({}^r \mathbf{p}^k), & u_y({}^r \mathbf{p}^k), & u_z({}^r \mathbf{p}^k) \end{bmatrix}^{\top},
    \end{equation}
    where $(u_x,u_y,u_z)$ are the components of the wind velocity along the Cartesian axes. This is the local wind model in Section~\ref{sec:propagation}, including its spatial variation, turbulence, and vortex effects. Since the onboard sensor is imperfect, the measured value is corrupted by additive Gaussian noise:
    \begin{equation}
        \mathbf{u}_{\text{meas}}({}^r \mathbf{p}^k) = \mathbf{u}_{\text{wind}}({}^r \mathbf{p}^k) + \boldsymbol{\eta}_w, \quad \boldsymbol{\eta}_w \sim \mathcal{N}(\mathbf{0}, \sigma_w^2 \mathbf{I}_3).
    \end{equation}
    The horizontal wind direction is characterized by the angle of $\mathbf{u}_{\text{meas}}$ relative to the $x$-axis in the $xy$-plane. This angle represents the prevailing advection direction of the plume.

\subsubsection{Local pollutant concentration estimate:} The UAV computes an estimated local concentration $\hat{C}({}^r \mathbf{p}^k)$ using the positions of nearby ground-truth plume particles $\{\mathbf{x}_i^{\text{plume}}\}_{i=1}^{N_{\text{GT}}}$ with a Gaussian kernel:
    \begin{equation}
        \hat{C}({}^r \mathbf{p}^k) = \sum_{i=1}^{N_{\text{GT}}} 
        \exp\Bigg(-\frac{\|{}^r \mathbf{p}^k - \mathbf{x}_i^{\text{plume}}\|^2}{2\sigma_s^2}\Bigg),
    \end{equation}
    where $\sigma_s$ denotes the effective sensing radius. Particles closer to the UAV contribute more strongly, reflecting spatial averaging of the sensor. To model measurement imperfections, additive Gaussian noise is included:
    \begin{equation}
        \hat{C}_\text{meas}({}^r \mathbf{p}^k) = \hat{C}({}^r \mathbf{p}^k) + \eta, \quad \eta \sim \mathcal{N}(0, \sigma_n^2).
    \end{equation}

\subsection{Local Concentration Gradient Estimation}

To navigate toward high-concentration regions, the UAV estimates the local concentration gradient using a six-point finite-difference stencil centered at its own position. For a small sensor offset $\delta>0$ and standard Cartesian unit vectors $\mathbf{e}_x$, $\mathbf{e}_y$, and $\mathbf{e}_z$, the UAV evaluates the noisy concentration estimate $\hat{C}_\text{meas}(\cdot)$ at six probe points displaced by $\pm\delta$ along each Cartesian axis,
\begin{equation}
{}^r\mathbf{p}^k_{\pm x} = {}^r\mathbf{p}^k \pm \delta\,\mathbf{e}_x, \;
{}^r\mathbf{p}^k_{\pm y} = {}^r\mathbf{p}^k \pm \delta\,\mathbf{e}_y, \;
{}^r\mathbf{p}^k_{\pm z} = {}^r\mathbf{p}^k \pm \delta\,\mathbf{e}_z,
\end{equation}
and forms the central-difference gradient estimate
\begin{equation}
\nabla \hat{C}({}^r \mathbf{p}^k) = 
\begin{bmatrix}
\dfrac{\hat{C}_\text{meas}({}^r\mathbf{p}^k_{+x}) - \hat{C}_\text{meas}({}^r\mathbf{p}^k_{-x})}{2\delta} \\[2mm]
\dfrac{\hat{C}_\text{meas}({}^r\mathbf{p}^k_{+y}) - \hat{C}_\text{meas}({}^r\mathbf{p}^k_{-y})}{2\delta} \\[2mm]
\dfrac{\hat{C}_\text{meas}({}^r\mathbf{p}^k_{+z}) - \hat{C}_\text{meas}({}^r\mathbf{p}^k_{-z})}{2\delta}
\end{bmatrix},
\label{eq:gradient_final}
\end{equation}
The resulting vector is normalized to a unit direction,
\begin{equation}
\overline{\nabla \hat{C}}({}^r \mathbf{p}^k) = \frac{\nabla \hat{C}({}^r \mathbf{p}^k)}{\|\nabla \hat{C}({}^r \mathbf{p}^k)\| + \epsilon},
\end{equation}
so that only the sensed direction of increasing concentration, not its magnitude, is passed to the guidance stage. This finite-difference formulation reflects a physically realizable six-sensor array mounted on the UAV, rather than an analytically differentiated closed-form gradient, and inherits the additive measurement noise of each individual probe.

\subsection{Wind-Adaptive Gradient-Based Surrogate Plume Propagation}

To incorporate atmospheric advection, the sensed local concentration gradients and wind measurements are combined to propagate the  surrogate plume points (or equivalently, SP). Each SP is moved by the measurements of the UAVs, weighted by $w_{jr}^k$, the inverse of its distance to each UAV, so that a nearby UAV has a stronger influence than a distant one:
\begin{equation}
\begin{aligned}
 \mathbf{q}_j^{k+1} = &\mathbf{q}_j^{k} + \Delta t \sum_{r=1}^{N_{\text{UAV}}} w_{jr}^k \left( \lambda_g \, \overline{\nabla \hat{C}}({}^r \mathbf{p}^k) + \mathbf{u}_{\text{meas}}({}^r \mathbf{p}^k) \right)\\
&+ \mathbf{r}_j^k + \boldsymbol{\xi}_j^k,
\label{eq:guidance_final}
\end{aligned}
\end{equation}
with
\vspace{-.1in}
\begin{equation}
 w_{jr}^k = \frac{\|\mathbf{q}_j^k - {}^r\mathbf{p}^k\|^{-1}}{\sum_{s=1}^{N_{\text{UAV}}} \|\mathbf{q}_j^k - {}^s\mathbf{p}^k\|^{-1}},
\end{equation}
where $\lambda_g$ weights the unit-norm gradient attraction against the wind advection, $\mathbf{r}_j^k$ is a short-range repulsion that pushes apart SP points closer than $d_{\min}$, and $\boldsymbol{\xi}_j^k$ is a wind-aligned stochastic diffusion that keeps the SP set spread out. The weights $w_{jr}^k$ sum to one over the UAVs, and thus each SP moves with an average of the measured gradients and winds, dominated by the UAVs nearest to it. For a single UAV, $w_{jr}^k = 1$ and \eqref{eq:guidance_final} reduces to a per-UAV update. Each UAV reports its sensed gradient and wind vectors to a central coordinator, which maintains the shared SP set.

\subsection{UAV Motion Tracking and Guidance}

The UAV's own position is not driven directly by \eqref{eq:guidance_final}. Instead, at each simulation step the UAV tracks the current D$^2$C goal point ${}^r\mathbf{q}_{\text{goal}}^k$ using the fixed-wing motion controller of Stage~B, so that the sensed gradient and wind information influence UAV motion only indirectly, through their effect on the SP distribution that defines ${}^r\mathbf{q}_{\text{goal}}^k$. This two-layer structure, combining sensing and SP propagation on one side with D$^2$C-based motion control on the other, forms the basis for integrating the sensing pipeline with the D$^2$C controller.

\section{D$^2$C-Based Multi-UAV Air Pollution Monitoring Algorithm}
\label{sec:D$^2$C_ai_uav}

This section presents the complete multi-UAV air pollution monitoring algorithm implemented and evaluated in this paper. At every discrete time step, the ground-truth plume propagates, and each UAV senses local concentration gradients and wind vectors. These sensor measurements are reported to a central coordinator, which updates a single shared surrogate plume (SP) set and its weight array. Subsequently, the D$^2$C controller (Stages A--C of Section~\ref{sec:preliminaries}) selects target points from this SP distribution, which each UAV then tracks using a fixed-wing motion controller.

Specifically, while individual UAVs independently acquire local sensor data, the SP set and its weights are maintained globally. The central coordinator aggregates the sensed gradient and wind vectors from all agents to update the shared reference distribution. 

\subsection{Algorithm Overview}

At each discrete sensing time step $k$, the overall system executes the following coordinated sequence of operations.

\subsubsection{Ground-Truth Plume Propagation:}

    The ground-truth plume particles are advanced by one sensing interval $\Delta t$ under the combined wind-advection, turbulence, L\'evy-flight, and vortex-induced motion model of Section~\ref{sec:propagation}.

\subsubsection{Local Sensing:}

UAV $r$ acquires raw onboard measurements consisting of a locally sensed wind vector ${}^r\mathbf{u}_{\text{wind}}^k$ (Section~\ref{sec:posterior_update}) and a locally sensed concentration gradient $\nabla \hat{C}_r^k$ obtained from the six-point finite-difference stencil in \eqref{eq:gradient_final}. These measurements reflect local atmospheric turbulence and vortex structures, and are subject to additive sensor noise. 

\subsubsection{Surrogate Plume (SP) Update:}

Using the sensed gradients and wind estimates collected from all UAVs, the central coordinator updates the surrogate plume (SP) set according to \eqref{eq:guidance_final}. The total number of SP remains fixed throughout the mission, with no particles added, discarded, or resampled via clustering.

\subsubsection{D$^2$C-Based Goal Selection and Weight Update:}

    Within the same sensing interval, at every UAV integration step, each UAV selects the nearest available SPs up to its per-step weight budget $\alpha_r^k$ and computes their weighted barycenter as its local goal point ${}^r\mathbf{q}_{\text{goal}}^k$ (Stage A); an appropriate control law is applied to each UAV to follow the designated goal (Stage B); and reduce the weight of some SPs for the coverage progress (Stage C). Since the local SP selection is sequential across UAVs within a step, a given SP point cannot be claimed by more than one UAV at the same instant, avoiding redundant area coverage.

\subsubsection{History Logging:}

    Ground-truth plume, SP, and UAV trajectory data are logged for subsequent analysis.

\subsection{Pseudocode Description}
The overall algorithm is summarized in the pseudocode below. To model a more realistic operational scenario, two distinct time intervals are introduced: $\Delta t_{\text{uav}}$ for the system control loop and $\Delta t_{\text{sense}}$ for the sensor sampling loop, which is the same as $\Delta t$ used in the previous sections. Typically $\Delta t_{\text{sense}} > \Delta t_{\text{uav}}$.
\begin{algorithm}[!h]
\caption{D$^2$C-Based Multi-UAV Air Pollution Monitoring}
\label{alg:D$^2$C_ai_uav}
\small
\begin{algorithmic}[1]
\State \textbf{Initialize:} UAV states ${}^r\mathbf{p}^0$ and SP weights $\boldsymbol{\beta}^0$ (shared SP pool), ground-truth plume, sensing interval $\Delta t_{\text{sense}}$, UAV control loop time $\Delta t_{\text{uav}}$, total mission time $T_{\text{total}}$, and per-step weight budget $\alpha = \Delta t_{\text{uav}}/(T_{\text{total}} N_{\text{UAV}})$
\For{$k = 0$ to $T_{\mathrm{total}} / \Delta t_{\text{sense}}$}
    \State \textbf{Propagate ground-truth plume} during $\Delta t_{\text{sense}}$
    \For{each UAV $r$}
        \State \textbf{Local sensing:} obtain $^r\mathbf{u}_{\text{wind}}^k$, $\nabla \hat{C}_r^k$
    \EndFor
    \State \textbf{SP update:} propagate $\mathbf{q}_j^{k+1}$ via \eqref{eq:guidance_final} and enforce $z \ge 0$
    \For{$i = 1$ to $\Delta t_{\text{sense}}/\Delta t_{\text{uav}}$}
        \For{each UAV $r$ (sequentially)}
            \State \textbf{Stage A:} compute ${}^r\mathbf{q}_{\text{goal}}^k$ from the available SPs
            \State \textbf{Stage C:} consume the SP weights by $\alpha$
        \EndFor
    \EndFor
    \For{each UAV $r$}
        \State \textbf{Stage B:} advance the UAV by one $\Delta t_{\text{uav}}$ step toward ${}^r\mathbf{q}_{\text{goal}}^k$
    \EndFor
    \State \textbf{Update} heading memory, and \textbf{log} plume, SP, and UAV trajectory history
\EndFor
\end{algorithmic}
\end{algorithm}

\subsection{Algorithmic Properties}

The algorithm is characterized by the following implementation features:
\begin{itemize}
    \item \textbf{Centralized SP Management:} A common set of SPs and their corresponding weights are maintained by the central coordinator. Individual UAVs use this shared information to determine their respective target locations and control actions.

    \item \textbf{Sensor-Driven Updates:} The SP distribution is modified using only the gradient and wind measurements collected by the UAVs. No ground-truth information regarding the actual plume distribution is used.

    \item \textbf{Unfiltered Measurements:} The measured gradient and wind vectors are directly incorporated into the algorithm, without additional filtering, state estimation, or learning-based correction.

    \item \textbf{Preserved SP Population:} The number of SPs is unchanged throughout the mission, with no particle resampling, clustering, insertion, or removal.
\end{itemize}
\section{Simulations}\label{sec:sim}

To evaluate the performance of the proposed centralized Density-Driven Control ($D^2C$) framework for real-time air pollution monitoring and tracking, a comprehensive simulation environment was implemented in MATLAB. The simulation integrates three-dimensional fixed-wing aircraft kinematics governed by bounded bank and pitch angles, operating within a dynamic pollutant field governed by the ground-truth propagation model in \eqref{eq:combined_model}. 

In this framework, individual UAVs independently acquire localized concentration and wind measurements, modeled as Gaussian-weighted sums of nearby ground-truth particles corrupted by additive sensor noise, and transmit them to a central coordinator. The coordinator maintains a single shared surrogate plume set and associated weight array to guide multi-agent coordination. 

The proposed $D^2C$ approach is benchmarked against two alternative monitoring strategies: 
\begin{enumerate}
    \item[(1)] \textbf{Prediction-Only (PRED):} A baseline utilizing an open-loop surrogate plume propagation model based on forecasted wind conditions that deviate from the ground-truth environment, isolating the accuracy achievable without real-time UAV feedback.
    \item[(2)] \textbf{Distributed Ergodic Coverage (HEDAC):} A heat-equation-driven ergodic coverage method, implemented following the distributed formulation of Mantovani \emph{et al.} \cite{mantovani2026distributed}, which combines Gaussian-process field estimation with distributed spatial coverage.
\end{enumerate}

\subsection{Simulation Setup and Scenario Design}
\label{sec:simulation_setup}

To distinguish between data-driven adaptation and open-loop forecasting, two distinct surrogate plume (SP) representations are maintained during the simulation:
\begin{itemize}
    \item \textbf{Green SP Set:} The reference distribution updated via the proposed $D^2C$ framework, which assimilates real-time UAV sensor measurements (concentration gradients and local wind vectors) to guide closed-loop adaptive monitoring.
    \item \textbf{Blue SP Set:} The reference distribution corresponding to the prediction-only (PRED) baseline, which propagates purely through open-loop forecast model assumptions without sensor feedback.
\end{itemize}

The ground-truth (GT) plume is initialized with $N_p=100$ particles drawn uniformly from the $50\times50\times50$~m release box $[50,100]\times[50,100]\times[100,150]$~m. The Green SP set, which serves as the reference distribution of D$^2$C, is drawn independently from the same box, so it knows the release region but not the exact GT particle positions. The Blue (prediction-only) SP set starts as a copy of the Green SP set and afterwards differs from it only in how it is propagated. Both SP sets therefore start from the same information. The Green SPs are updated with the UAV measurements, and the Blue SPs follow the forecast wind alone.

\subsubsection{Fixed-Wing UAV Dynamics}

Each UAV is modeled as a 3D fixed-wing aircraft equipped with bounded bank and pitch angles. The discrete-time kinematics for UAV $r$ at time step $k$ are given in \cite{kai2019unified} by
\begin{align}
{}^r x^{k+1} &= {}^r x^k + \Delta t_{\text{uav}} \cdot \, {}^r v \cos {}^r \theta^k \cos {}^r \psi^k, \\
{}^r y^{k+1} &= {}^r y^k + \Delta t_{\text{uav}} \cdot \, {}^r v \cos {}^r \theta^k \sin {}^r \psi^k, \\
{}^r z^{k+1} &= {}^r z^k + \Delta t_{\text{uav}} \cdot \, {}^r v \sin {}^r \theta^k, \\
{}^r \psi^{k+1} &= {}^r \psi^k + \Delta t_{\text{uav}} \cdot \, {}^r \psi_{\text{rate}}^k, \\
{}^r \theta^{k+1} &= {}^r \theta^k + \Delta t_{\text{uav}} \cdot \, {}^r \theta_{\text{rate}}^k,
\end{align}
where ${}^r v$ is the UAV airspeed, ${}^r \psi$ the yaw (heading), ${}^r \theta$ the pitch, and ${}^r \psi_{\text{rate}}^k$ and ${}^r \theta_{\text{rate}}^k$ denote the yaw and pitch angular rates, respectively. The yaw rate follows from a coordinated turn with a commanded bank angle ${}^r\phi^k$, and the pitch rate is a saturated proportional command:
\begin{align}
{}^r\phi^k &= \text{sat}_\phi\big(k_{\psi\to\phi} \, {}^r \psi_{\text{error}}^k \big), \quad |{}^r\phi^k| \le \phi_{\max}, \\
{}^r \psi_{\text{rate}}^k &= \frac{g \tan {}^r\phi^k}{{}^r v}, \\
{}^r \theta_{\text{rate}}^k &= \text{sat}_\theta\big(k_\theta \, {}^r \theta_{\text{error}}^k \big), \quad |{}^r \theta^{k+1}| \le \theta_{\max},
\end{align}
where $g$ is the gravitational acceleration, with the yaw and pitch errors defined by
\begin{align}
{}^r \psi_{\text{error}}^k &= {}^r \psi_{\text{desired}}^k - {}^r \psi^k, \quad {}^r \psi_{\text{error}}^k \in [-\pi, \pi], \\
{}^r \theta_{\text{error}}^k &= {}^r \theta_{\text{desired}}^k - {}^r \theta^k, \quad {}^r \theta_{\text{error}}^k \in \left[-\frac{\pi}{2}, \frac{\pi}{2}\right].
\end{align}
The desired yaw and pitch, ${}^r \psi_{\text{desired}}^k$ and ${}^r \theta_{\text{desired}}^k$, are the azimuth and elevation of the direction from the UAV to its goal point ${}^r\mathbf{q}_{\text{goal}}^k$. The symbols $\text{sat}_\phi(\cdot)$ and $\text{sat}_\theta(\cdot)$ are saturation functions, $k_{\psi\to\phi}$ and $k_\theta$ are proportional gains, and $\phi_{\max}$ and $\theta_{\max}$ denote the maximum bank and pitch angles. All parameters and their values related to UAVs and sensing are listed in Table \ref{tab:uav_sensing}.
The pollutant samples are propagated and sensed every $\Delta t_\text{sense}=1$~s, and the UAV kinematics are integrated every $\Delta t_\text{uav}=0.1$~s, since the pollutant field evolves slowly compared with the UAV dynamics.

\begin{table}[t]
\centering
\begin{threeparttable} 
\caption{UAV and Sensing Parameters}
\label{tab:uav_sensing}
\renewcommand{\arraystretch}{1.1}
\setlength{\tabcolsep}{3.8pt}
\footnotesize
\begin{tabular}{llll}
\hline
\textbf{Parameter} & \textbf{Symbol} & \textbf{Value} & \textbf{Unit} \\ \hline
UAV integration timestep & $\Delta t_{\text{uav}}$ & 0.1 & s\\
Sensing/propagation interval & $\Delta t_{\text{sense}}$ & 1 & s\\
 UAV total operation time &  $T_{\text{total}}$ &  1 or 5 &  hr \\
\# of UAVs & $N_{\text{UAV}}$ & 3 & – \\
\# of GT plume particles & $N_{GT}$ & 100 & – \\
 \# of Green / Blue SP points &  $N_{\text{REF}}$ / $N_{\text{PRED}}$ &  100 &  – \\
 Agent's weight per step &  $\alpha_r^k$ &  $\footnotesize \frac{\Delta t_{\text{uav}}}{ (T_{\text{total}} N_{\text{UAV}})}$ &  – \\
 Coordination &  – &  centralized &  – \\
 Yaw-to-bank gain &  $k_{\psi\to\phi}$ &  5 &  – \\
Pitch gain & $k_\theta$ & 3 & – \\
 Airspeed (constant) &  $v$ &  20 &  m/s \\
 Maximum bank angle &  $\phi_{\max}$ &  35 &  deg \\
Maximum pitch angle & $\theta_{\max}$ & 20 & deg \\
Initial position range & – & $[0,50]^3$ {\footnotesize (uniform)} & m \\ \hline
Gaussian kernel width & $\sigma_s$ & 15 & m \\
Finite-diff. sensor offset & $\delta$ & 1.0 & m \\
Concentration noise std. & $\sigma_n$ & 0.05 & – \\
Wind measurement noise std. & $\sigma_w$ & 0.3 & m/s \\
 Gradient-attraction gain &  $\lambda_g$ &  0.2 &  m/s \\
Green SP repulsion distance & $d_{\min}$ & 35 & m \\
 Green SP diffusion std. &  – &  20 &  m \\
 Blue SP diffusion std. &  – &  15 &  m \\ \hline
\end{tabular}
\begin{tablenotes} 
\item \footnotesize{\textit{Note:} The simulations employ a group of three UAVs, each independently initialized with a uniformly random position in $[0,50]\times[0,50]\times[0,50]$\,m and a random heading.} 
\end{tablenotes}
\end{threeparttable} 
\end{table}

\subsubsection{Pollutant Field Representation and Environmental Setup}

Table~\ref{tab:env_params} lists the parameters of the GT propagation model and the nominal (forecast) parameters of the PRED set for the mild and strong wind scenarios. In both scenarios, the forecast wind direction is biased from the initial GT direction ($15^\circ$ for mild wind and $30^\circ$ for strong wind), the forecast speed is lower than the GT speed, and the GT direction also follows a random walk that the forecast does not capture. The REF set corrects these errors with the UAV measurements.

The pollutant field is represented by three sample sets. The first is the GT sample set,
\[
\mathcal{S}_{\text{GT}} = \{ \mathbf{x}_j^{\text{GT}} \}_{j=1}^{N_{\text{GT}}}, \quad \mathbf{x}_j^{\text{GT}} \in \mathbb{R}^3,
\]
which evolves according to the GT environmental parameters in Table~\ref{tab:env_params}.

The second is the PRED set, which evolves under the nominal parameters without UAV measurements and therefore drifts from the GT over time. The third is the REF set,
\[
 \mathcal{S}_{\text{REF}} = \{ \mathbf{q}_j^k \}_{j=1}^{N_{\text{REF}}}, \quad \boldsymbol{\beta}^k = \{ \beta_j^k \}_{j=1}^{N_{\text{REF}}},
\]
which is updated at every time step from the UAV measurements by \eqref{eq:guidance_final}. The GT sample positions enter only through these Gaussian-weighted measurements with noises and are never available to the UAVs.

\begin{table}[t]
\caption{Environmental Parameters for Mild and Strong Wind Scenarios}
\label{tab:env_params}
\centering
\scriptsize
\setlength{\tabcolsep}{3.2pt}
\resizebox{\columnwidth}{!}{%
\begin{tabular}{lcccc}
\toprule
\multirow{2}{*}{Parameter} & \multicolumn{2}{c}{Mild Wind} & \multicolumn{2}{c}{Strong Wind} \\
\cmidrule(lr){2-3} \cmidrule(lr){4-5}
&  GT &  Pred. &  GT &  Pred. \\
\midrule
 Propagation interval (s) &  1 &  1 &  1 &  1 \\
 Simulation time (hr) &  1, 5 &  1, 5 &  1, 5 &  1, 5 \\
 \# of samples &  100 &  100 &  100 &  100 \\
Mean wind speed (m/s) & 3 & 2.5 & 8 & 7 \\
 Initial wind dir. &  $30^\circ$ &  $45^\circ$ &  $30^\circ$  &  $60^\circ$  \\
 Dir. random walk std. (per s) &  $0.75^\circ$ &  -- &  $0.75^\circ$ &  -- \\
 Vertical wind component &  0.05 &  0.05 &  0.05 &  0.05 \\
 Turbulence std. (m/$\sqrt{\text{s}}$) &  20 &  -- &  20 &  -- \\
 Cross-wind turbulence factor &  0.1 &  -- &  0.1 &  -- \\
 Isotropic diffusion std. (m) &  -- &  15 &  -- &  15 \\
 Vertical shear coefficient &  0.02 &  -- &  0.02 &  -- \\
\midrule
\multirow{3}{*}{Vortex field}
&  No. vortices = 30 &  -- &  No. vortices = 30 &  -- \\
&  Strength = 5 m/s &  -- &  Strength = 5 m/s &  -- \\
&  Radius 50--300 m &  -- &  Radius 50--300 m &  -- \\
\midrule
 L\'evy flight (Cauchy) &  Scale = $10^{-5}$ m &  -- &  Scale = $10^{-5}$ m &  -- \\
\bottomrule
\end{tabular}%
}
\par\smallskip
\footnotesize\textit{Note:} The vortex centers are drawn uniformly over the region the plume can reach during the mission. The prediction (Pred.) set is advected along the fixed nominal direction with isotropic diffusion only; it models neither the direction drift, the turbulence, nor the vortices.
\end{table}

\subsubsection{Baseline: Distributed Ergodic Coverage with GP Estimation (HEDAC)}
\label{sec:hedac_baseline}

To provide a rigorous benchmark, we also compare against a Heat-Equation-Driven Area Coverage (HEDAC) method, implemented following the distributed formulation of Mantovani \emph{et al.}~\cite{mantovani2026distributed}. Rather than utilizing a surrogate plume model, HEDAC relies on a non-parametric approach as follows:
\begin{enumerate}[label=\roman*), leftmargin=*, labelsep=5pt]
    \item Gaussian Process (GP) Field Estimation: Each UAV fits a local Gaussian process to its concentration measurements, discarding outdated samples via a temporal decay filter to track time-varying pollutant concentrations.
    \item Ergodic Search Trajectory Generation: A target density is derived from the GP posterior, and a heat-equation potential field is solved over the workspace to compute search trajectories that balance exploration and high-density region coverage.
\end{enumerate}

To ensure a fair and consistent evaluation, the HEDAC implementation utilizes the exact same fixed-wing UAV kinematics, airspeed constraints, and sensor models as our proposed framework, with its internal parameters configured according to the guidelines in~\cite{mantovani2026distributed}.

Unlike D$^2$C, HEDAC requires a monitoring domain, and the choice of this domain strongly affects its performance on a moving plume. Two domain choices are therefore evaluated:
\begin{itemize}
    \item \textbf{Fixed Domain:} The fixed domain is a static bounding box designed to contain the entire plume trajectory. It extends beyond the initial release volume by the maximum drift calculated using the ground-truth mean wind speed over the total mission duration $T_{\text{total}}$. Although this idealized setup assumes a priori knowledge of the true wind and ensures that the plume never leaves the domain, it intentionally serves as an upper-bound baseline that heavily favors the baseline method. To maintain computational tractability, the HEDAC potential equations are solved using a grid cell size of $400\,\text{m} \times 400\,\text{m} \times 200\,\text{m}$ for 1\,h missions, transitioning to coarser cell sizes ($800\,\text{m} \times 800\,\text{m} \times 400\,\text{m}$ for mild wind; $2100\,\text{m} \times 2100\,\text{m} \times 1050\,\text{m}$ for strong wind) during 5\,h missions.

    \item \textbf{Moving Domain:} The moving domain is constructed dynamically from HEDAC's own Gaussian process (GP) posterior estimates. Every 10\,s, each UAV draws 100 sample points from its GP posterior (incorporating the mean plus one standard deviation to account for uncertainty). The domain boundary is then defined by the outer envelope enclosing these pooled points, expanded outward by a margin of one grid cell (corresponding to a 100\,m spatial resolution). If all UAV maps lose track of the plume, the domain boundary is held static. At $t=0$, the domain coincides with the initial release volume. This update mechanism represents our extension of HEDAC, as the original formulation in \cite{mantovani2026distributed} assumes a fixed domain. Because HEDAC lacks wind information, the resulting domain expands symmetrically around the locally sensed locations and remains anchored near the release point even as the plume drifts away.
\end{itemize}

For quantitative evaluation under both domain configurations, 100 points are drawn per UAV from its GP posterior mean over the domain (with negative values and values below 5\% of the maximum set to zero). The resulting 300 points from the three UAVs are pooled and compared against the ground-truth samples, with the score averaged over five random draws to account for sampling variance. If a UAV's posterior mass falls below 1\% of its initial value, its map is considered lost, and points are drawn uniformly over the domain.

\subsubsection{Evaluation Metric}
The accuracy of the D$^2$C REF set, the PRED set, and the HEDAC maps is measured by their distance to the GT sample distribution. Optimal transport metrics quantify the dissimilarity between two empirical distributions, where smaller values indicate a closer match. Since computing the exact Wasserstein distance over extended mission durations (1\,h or 5\,h) is computationally prohibitive, we evaluate the discrepancy between the estimated and ground-truth plume distributions every 60\,s using the \emph{Sinkhorn distance}~\cite{cuturi2013sinkhorn}. As an entropy-regularized approximation of the optimal transport distance, the Sinkhorn distance preserves the full 3D transport geometry while remaining computationally tractable for our sample size, yielding a robust spatial dissimilarity metric in meters.

\subsection{Result and Analysis}
Four scenarios were simulated by combining two wind conditions (mild and strong, Table~\ref{tab:env_params}) with two mission lengths (1 and 5 hours). In each scenario, D$^2$C is compared with PRED and with HEDAC on the fixed and the moving domain. A random seed fixes the GT wind-direction random walk, the turbulence, the vortices, and the sensor noise, and separate random streams give all methods the same GT plume for a given seed. Trajectories are first shown for one seed, followed by a Monte Carlo evaluation over ten seeds and a robustness study with respect to the forecast wind direction.

\begin{figure*}[!t]
    \centering
    \includegraphics[width=\linewidth]{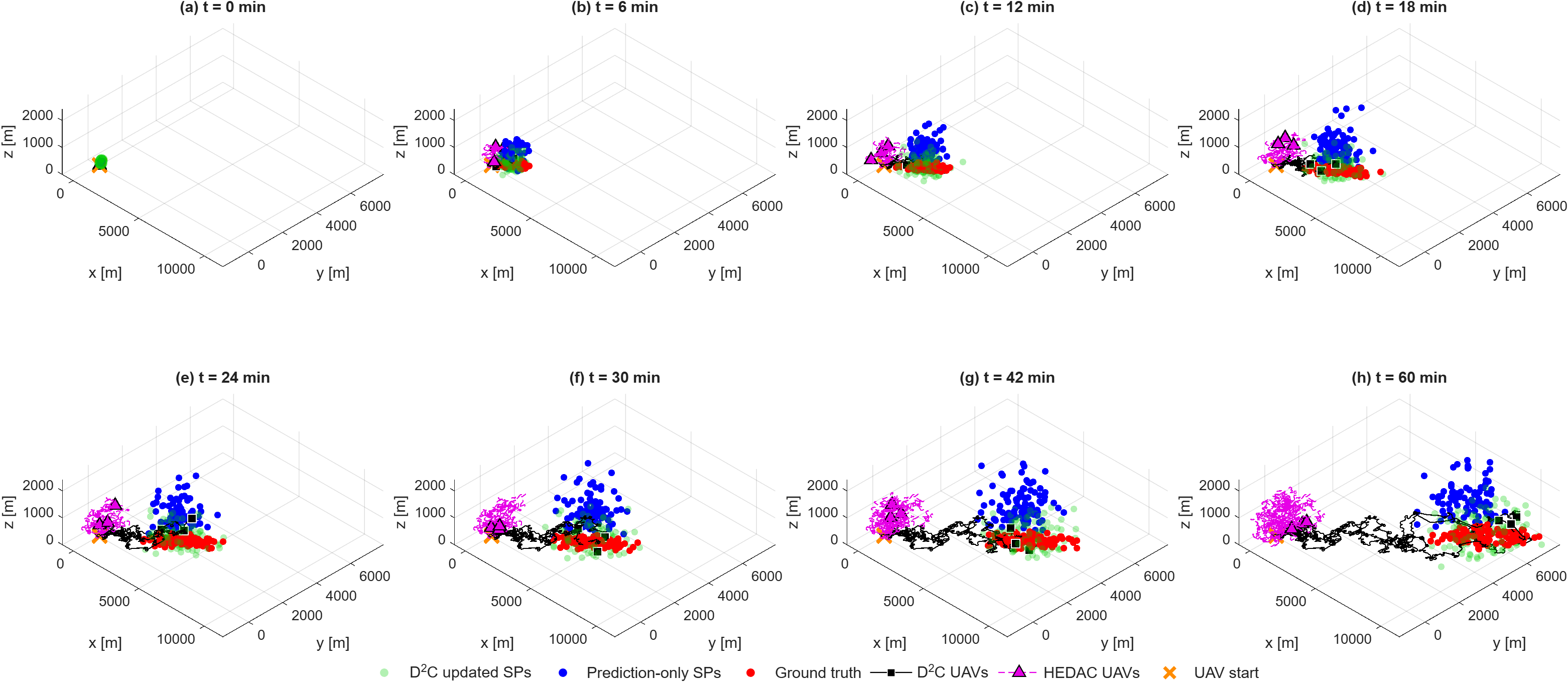}
    \caption{Three-UAV trajectories under \textbf{mild wind} over 1 hour, D$^2$C and HEDAC with moving domain. Red: GT samples; blue: prediction-only SPs; green: D$^2$C updated SPs; black solid lines with squares: D$^2$C UAVs; magenta dashed lines with black-outlined triangles: HEDAC UAVs; orange cross: UAV start position.}
    \label{fig:mild_1h_traj}
\end{figure*}

\begin{figure*}[!t]
    \centering
    \includegraphics[width=\linewidth]{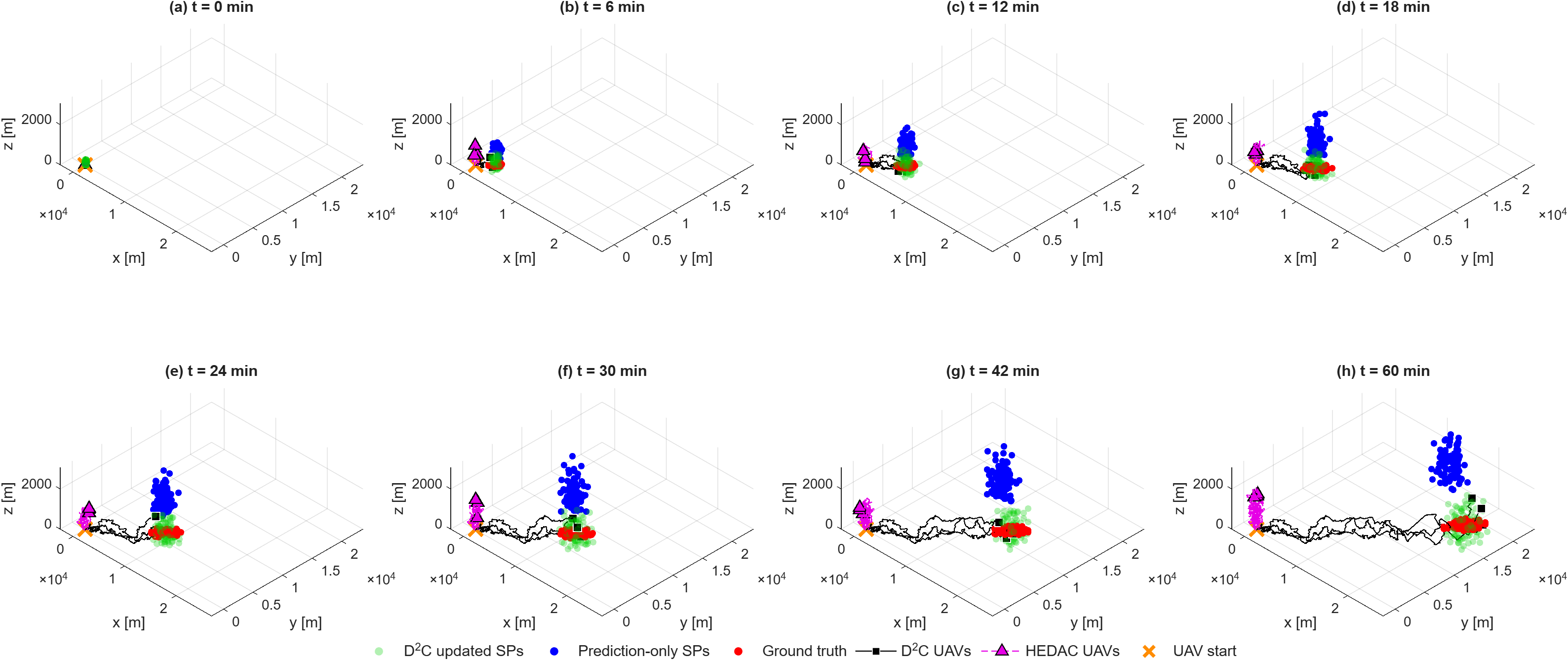}
    \caption{Three-UAV trajectories under \textbf{strong wind} over 1 hour, D$^2$C and HEDAC on the moving domain. Markers as in Figure~\ref{fig:mild_1h_traj}.}
    \label{fig:strong_1h_traj}
\end{figure*}

\subsubsection{Trajectories in the Short-Term Missions}

Figures~\ref{fig:mild_1h_traj} and \ref{fig:strong_1h_traj} illustrate the 1-hour missions under mild and strong wind conditions, comparing D$^2$C and HEDAC with the moving domain. While the GT plume drifts significantly over time, the proposed D$^2$C framework maintains continuous and robust tracking. Specifically, driven by measured wind advection and concentration gradient attraction, the green reference sample points (REF SPs) remain consistently and accurately overlapped with the red GT samples throughout the entire mission, enabling the UAVs to stay precisely within the plume core.

In contrast, HEDAC lacks explicit wind information. Under the moving domain configuration, its boundary remains anchored near the release point and fails to extend beyond roughly 2.3\,km, missing the rapidly drifting plume (which travels about 10\,km under mild wind and 28\,km under strong wind).

\begin{figure*}[!t]
 
    \centering
    \includegraphics[width=\linewidth]{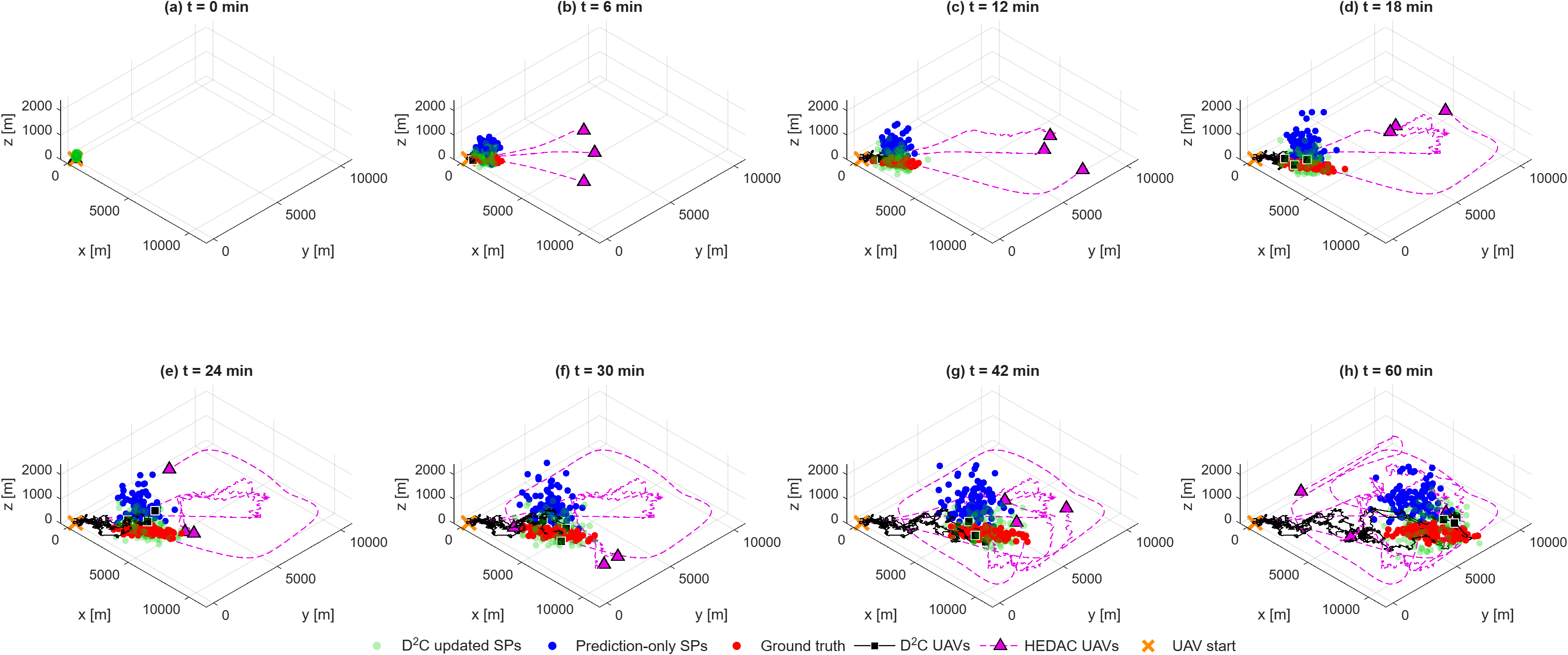}
    \caption{Three-UAV trajectories under \textbf{mild wind} over 1 hour, D$^2$C and HEDAC on the fixed domain. Markers as in Figure~\ref{fig:mild_1h_traj}.}
    \label{fig:mild_1h_fixed}
\end{figure*}

Figure~\ref{fig:mild_1h_fixed} presents the same mild-wind mission with HEDAC configured on a large fixed domain. Within this static bounding box, the drifting plume occupies only a small fraction of the total volume. Because HEDAC's Gaussian process is not informed by wind advection, the ergodic exploration disperses search efforts across the entire domain rather than concentrating on the narrow plume trajectory, resulting in only intermittent encounters and a map dominated by zero-concentration readings.

\begin{figure*}[!t]
 
    \centering
    \includegraphics[width=\linewidth]{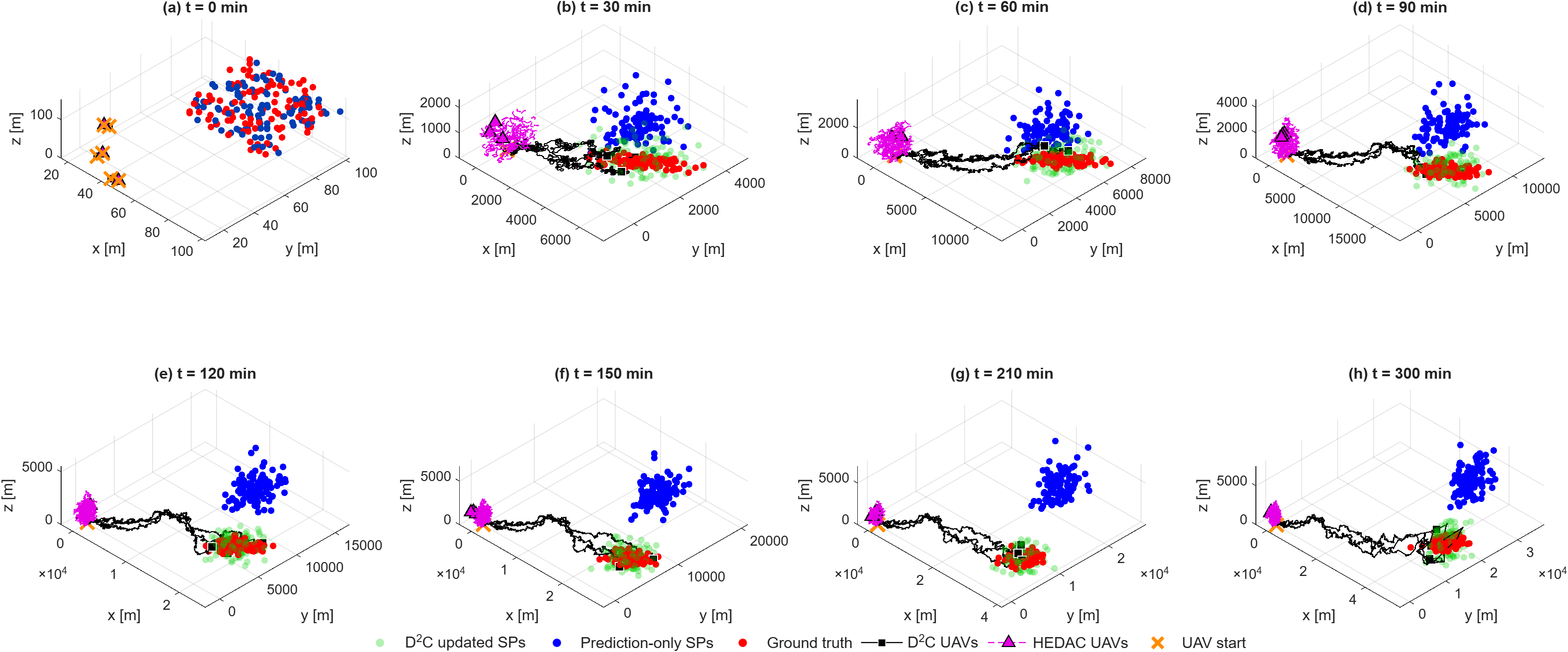}
    \caption{Three-UAV trajectories under \textbf{mild wind} over 5 hours, D$^2$C and HEDAC on the moving domain. Markers as in Figure~\ref{fig:mild_1h_traj}.}
    \label{fig:mild_5h_traj}
\end{figure*}

\begin{figure*}[!t]
 
    \centering
    \includegraphics[width=\linewidth]{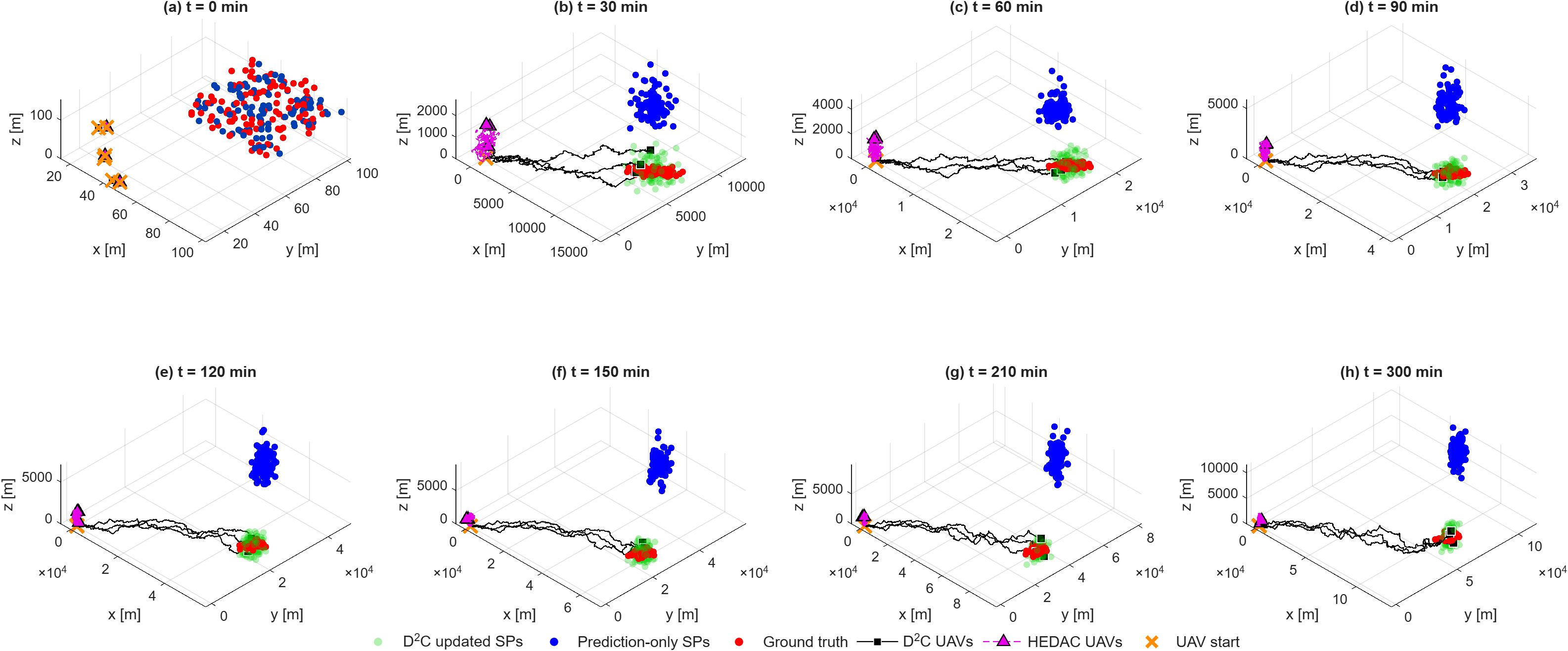}
    \caption{Three-UAV trajectories under \textbf{strong wind} over 5 hours, D$^2$C and HEDAC on the moving domain. Markers as in Figure~\ref{fig:mild_1h_traj}.}
    \label{fig:strong_5h_traj}
\end{figure*}

\subsubsection{Trajectories in the Long-Term Missions}

Figures~\ref{fig:mild_5h_traj} and \ref{fig:strong_5h_traj} illustrate the 5-hour extended missions, over which the cumulative effects of continuous wind-driven advection become dominant. As the plume travels vast distances, approximately 50\,km under mild wind and about 130\,km under strong wind, the open-loop forecast progressively deviates, ending up tens of kilometers away from the actual plume location. 

Despite this substantial drift over extended timescales, the proposed $D^2C$ framework demonstrates remarkable long-term stability and robustness. Throughout both mild and strong wind 5-hour missions, the $D^2C$ UAVs successfully remain with the plume from start to finish, with the green REF SPs continuously and accurately enclosing the red GT samples. 

In sharp contrast, HEDAC with the moving domain remains severely restricted near the release point (within roughly 2.6\,km), causing the UAVs to lose contact with the GT plume almost immediately after the mission begins. HEDAC trajectories on the fixed domain are omitted from the visual plots. Because the bounding box must span the entire plume trajectory (approximately $50\,\text{km}$ under mild wind and $130\,\text{km}$ under strong wind), the resulting scale causes both the plume and the UAV paths to occupy only a negligible fraction of the figure. However, its quantitative performance is included in the subsequent evaluation results using the coarser grid cell sizes specified in Section~\ref{sec:sim}-A.

\begin{figure*}[!h]
 
    \centering
    \includegraphics[width=0.8\linewidth]{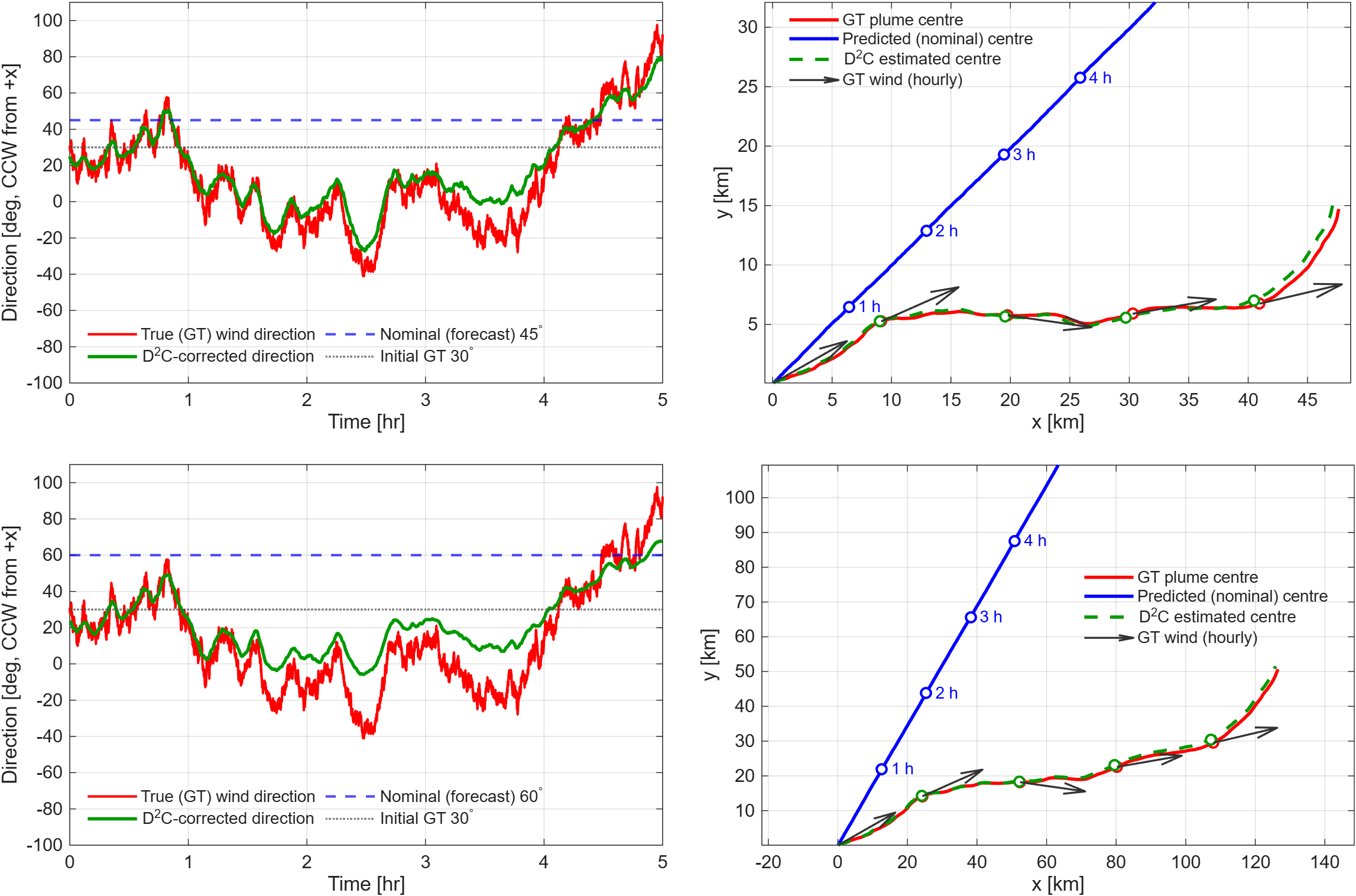}
    \caption{Wind model over 5 hours for mild (top) and strong (bottom) wind. Left: true (GT) wind direction, nominal (forecast) direction, and D$^2$C-corrected direction (heading of the REF SP centroid). Right: top view of the GT plume center, the predicted (nominal) center, and the D$^2$C estimated center, with hourly markers and the GT wind direction every hour.}
    \label{fig:wind_model}
\end{figure*}
\subsubsection{Wind Estimation and Plume Center Tracking}

Figure~\ref{fig:wind_model} shows why the forecast fails over long horizons. In the left panels, the GT wind direction drifts between about $-40^\circ$ and $+95^\circ$ over 5 hours, while the forecast direction stays constant. The D$^2$C-corrected direction, the heading of the REF SP centroid over a 5-minute window, follows the GT direction closely under mild wind. Under strong wind, it deviates by $10^\circ$--$30^\circ$ between 2 and 4 hours, because the faster plume spreads more along the wind and the centroid heading responds more slowly. In the right panels, the wind-driven trajectory is shown in the $\mathsf{xy}$-plane. While the forecast center moves along a straight line and ends far from the GT center, the estimated center of D$^2$C remains tightly locked onto the GT path in both scenarios (1 hour and 5 hours). Consequently, any directional deviation even under strong wind does not accumulate into a structural tracking error, as the measured concentration gradient continuously pulls and re-anchors the REF SPs back to the plume core.

\begin{figure*}[!h]
 
    \centering
    \includegraphics[width=0.8\linewidth]{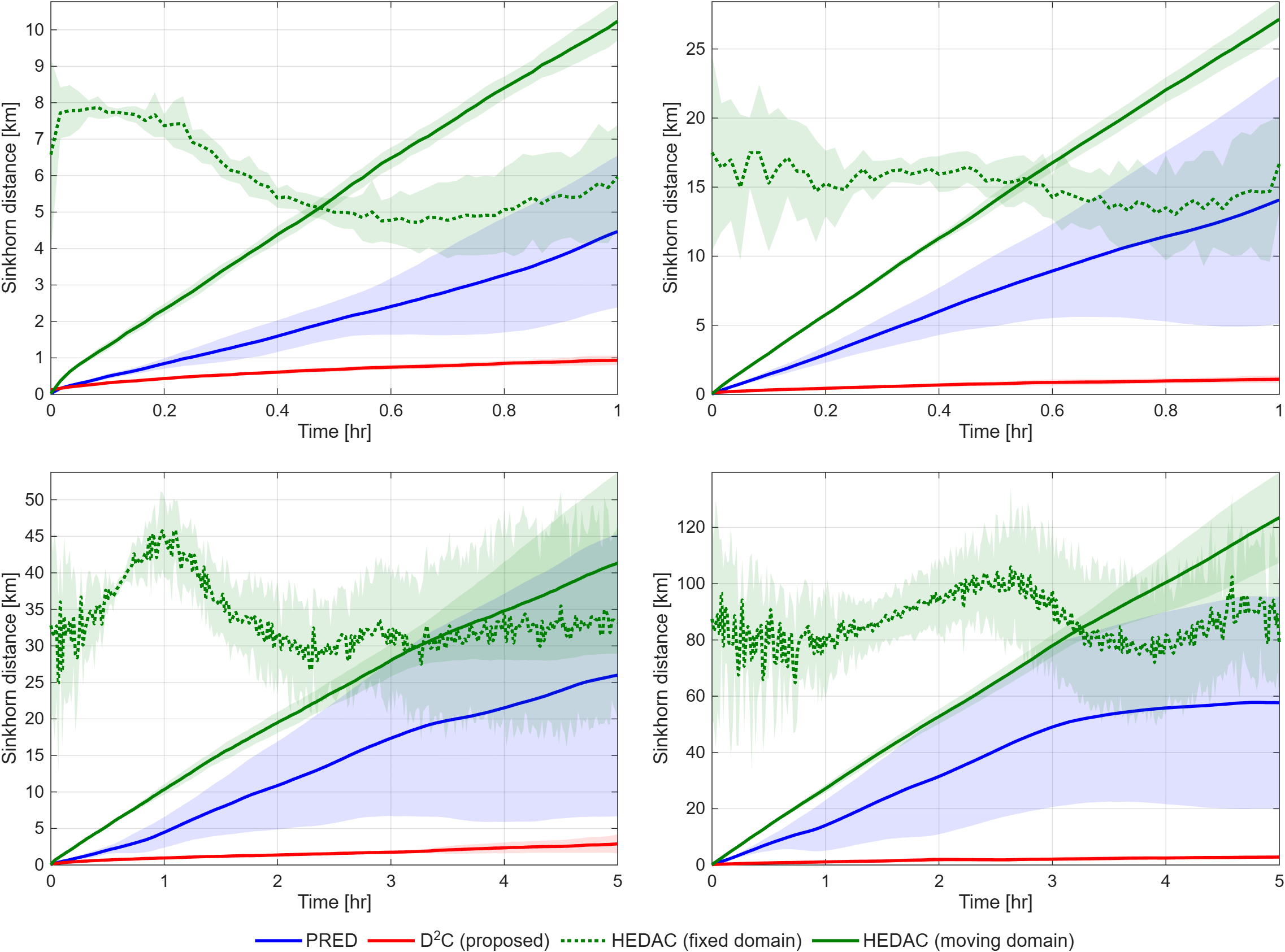}
    \caption{Sinkhorn distance to the GT samples over time, mean (lines) $\pm$ one standard deviation (shaded) over ten seeds. Top: 1-hour missions; bottom: 5-hour missions; left: mild wind; right: strong wind.}
    \label{fig:sinkhorn}
\end{figure*}

\begin{table}[!h]
 
\caption{Final Sinkhorn Distance to GT (km), Mean $\pm$ Std. over Ten Seeds}
\label{tab:mc_results}
\centering
\scriptsize
\setlength{\tabcolsep}{2.5pt}
\resizebox{\columnwidth}{!}{%
\begin{tabular}{lcccc}
\toprule
Scenario & D$^2$C & PRED & HEDAC fixed & HEDAC moving \\
\midrule
Mild, 1 h & $\mathbf{0.93 \pm 0.13}$ & $4.47 \pm 2.08$ & $5.96 \pm 1.42$ & $10.24 \pm 0.53$ \\
Strong, 1 h & $\mathbf{1.10 \pm 0.27}$ & $14.07 \pm 8.97$ & $16.67 \pm 3.42$ & $27.15 \pm 1.28$ \\
Mild, 5 h & $\mathbf{2.89 \pm 1.31}$ & $25.99 \pm 19.33$ & $33.27 \pm 13.34$ & $41.34 \pm 12.42$ \\
Strong, 5 h & $\mathbf{2.82 \pm 0.59}$ & $57.69 \pm 37.64$ & $78.72 \pm 19.22$ & $123.48 \pm 16.10$ \\
\bottomrule
\end{tabular}%
}
\end{table}
\subsubsection{Quantitative Performance Evaluation via Sinkhorn Distance}

The monitoring and tracking accuracy was quantified with the Sinkhorn distance in a Monte Carlo study with ten seeds per scenario (80 simulations in total). Figure~\ref{fig:sinkhorn} shows the mean and standard deviation of the distance over time, and Table~\ref{tab:mc_results} lists its final values.

D$^2$C achieves the lowest Sinkhorn distance in all four scenarios and at all times (Figure~\ref{fig:sinkhorn}). Its final distance is 0.93, 1.10, 2.89, and 2.82\,km in the mild 1\,h, strong 1\,h, mild 5\,h, and strong 5\,h scenarios, about 5, 13, 9, and 20 times lower than that of PRED (Table~\ref{tab:mc_results}). The PRED error grows almost linearly with time, at a rate set by the mismatch between the true and forecast winds, while the D$^2$C error grows slowly because the measured wind continually corrects the REF SPs. Across seeds, the standard deviation of the final D$^2$C distance is at most 1.31\,km, compared with up to 37.6\,km for PRED, whose error depends on how far the GT direction drifts from the forecast in each seed.

HEDAC on the fixed domain is less accurate than PRED in all scenarios, because its map spreads over the whole box, whose size grows with the wind speed and mission length. HEDAC on the moving domain yields the largest final distances, from 10.24\,km (mild, 1\,h) to 123.48\,km (strong, 5\,h), and its distance grows almost linearly with time (Figure~\ref{fig:sinkhorn}). Its domain stays within a few kilometers of the release point, and thus the final distance is close to the distance traveled by the plume. For the same reason, its standard deviation across seeds is smaller than that of PRED in every scenario: the estimate barely depends on where the plume drifts. A GP can only estimate the field inside its domain, and without wind information HEDAC cannot move this domain along with the plume.

\subsubsection{Robustness to the Forecast Wind Direction}

\begin{figure}[!h]
 
    \centering
    \includegraphics[width=0.82\linewidth]{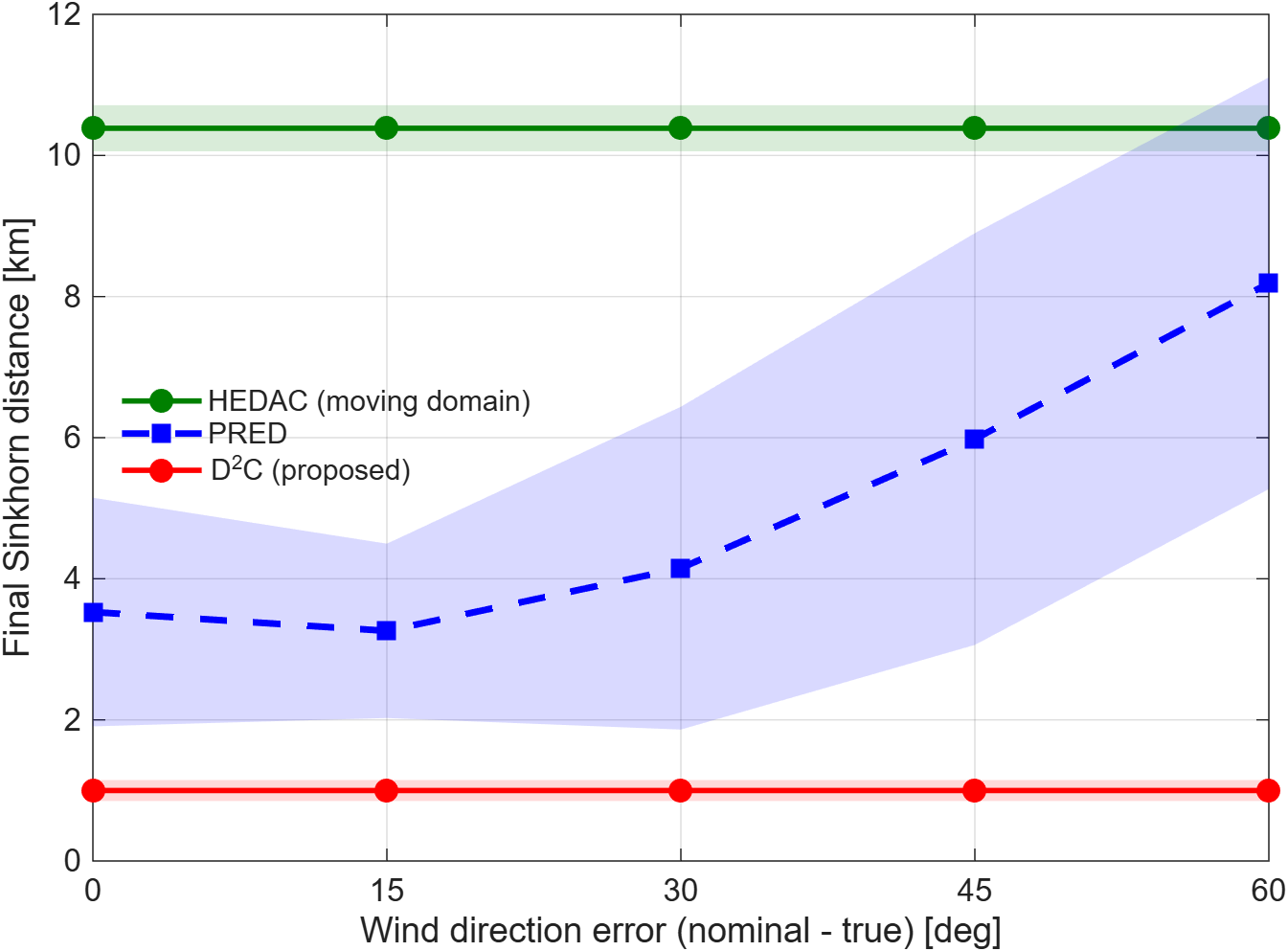}
    \caption{Final Sinkhorn distance after the 1-hour mild-wind mission versus the forecast wind direction error, mean $\pm$ one standard deviation over five seeds per offset.}
    \label{fig:sweep}
\end{figure}

To systematically evaluate how forecast inaccuracies impact long-term tracking performance, we examine the sensitivity of the system to errors in the wind direction as described in Figure \ref{fig:sweep}. 
Specifically, the mild-wind 1-hour mission was repeated with the forecast direction intentionally offset from the true initial direction by $0^\circ$, $15^\circ$, $30^\circ$, $45^\circ$, and $60^\circ$ (at a forecast speed of 2.5\,m/s), using five random seeds per offset.
Since PRED relies directly on the forecast, its accuracy is inherently susceptible to these offset errors. Figure~\ref{fig:sweep} shows the final Sinkhorn distance across the evaluated offsets. The error of PRED increases from about 3.5\,km to 8.2\,km over this range. Even at $0^\circ$ offset, its error is about 3.5 times that of D$^2$C, primarily due to the speed mismatch and the random walk inherent in the GT direction. In contrast, neither D$^2$C nor HEDAC utilizes the forecast direction, meaning their performance remains invariant to the offset: the D$^2$C error stably maintains around 1.0\,km as its REF SPs are propagated via measured wind, whereas the HEDAC error stays at about 10.4\,km.

Overall, the forecast alone drifts far from the GT, and a coverage method without wind information cannot follow this drift: a fixed domain large enough to contain the plume dilutes the estimate, and a domain built from the GP estimate stays near the release point. Satellite observations and ground-based stations cannot measure continuously at fine temporal resolution, and D$^2$C provides accurate and repeatable tracking of the pollutant field between their observations.


\section{Conclusion}
\label{sec:conclusion}

This paper presented an extension of the Density-Driven Control (D$^2$C) framework for centralized multi-UAV coordination in real-time air pollution monitoring and tracking. By leveraging Optimal Transport theory, the proposed approach dynamically guides UAVs toward regions with higher pollutant densities while continuously updating a probabilistic reference distribution using both predictive propagation and onboard measurements. A sensor-driven surrogate plume update mechanism, combining locally sensed gradient attraction, wind advection, inter-sample repulsion, and stochastic diffusion, enables UAVs to maintain accurate spatio-temporal estimates of the pollutant field, even under time-varying and spatially heterogeneous conditions.

Simulation results across 1-hour and 5-hour missions under mild and strong wind conditions highlight the limitations of existing approaches: both pure open-loop prediction and HEDAC fail to robustly track the moving source due to their lack of real-time wind-driven adaptation. In contrast, the proposed $D^2C$ approach achieves accurate and responsive reconstruction of the pollutant distribution over extended periods. Its final Sinkhorn distance to the ground truth remained below 3\,km across all scenarios, achieving a 5 to 20 times reduction compared to the prediction-only reference, with an order-of-magnitude lower variability across seeds and strong robustness against forecast wind errors. 

These results demonstrate that by dynamically shaping the reference distribution via wind-aware predictive transport and continuously re-anchoring it with real-time sensor measurements, D$^2$C effectively directs UAVs to systematically explore and cover the true spatial extent of the drifting plume where open-loop and non-adaptive baselines fail.

\bibliographystyle{ieeetr}
\bibliography{reference_arxiv}

\end{document}